\documentclass[final]{agujournal2019}
\usepackage{amsmath}
\usepackage{amsfonts}
\usepackage{booktabs}

\journalname{Journal of Advances in Modeling Earth Systems (JAMES)}

\begin{document}

%
%



\title{Long-Range Distillation: Distilling 10,000 Years of Simulated Climate into Long Timestep AI Weather Models}

%
%




\authors{Scott A. Martin\affil{1,2}\thanks{Research conducted during an internship at NVIDIA Research}, Noah Brenowitz\affil{1}, Dale Durran\affil{1,3}, Michael Pritchard\affil{1}}
\affiliation{1}{NVIDIA Research, Santa Clara, CA, USA}
\affiliation{2}{School of Oceanography, University of Washington, Seattle, WA, USA}
\affiliation{3}{Department of Atmospheric \& Climate Science, University of Washington, Seattle, WA, USA}






\correspondingauthor{Scott A. Martin}{scomartin@nvidia.com}

\begin{keypoints}
\item We replace long autoregressive ensemble rollouts for long-range weather forecasting with a single–timestep probabilistic prediction model.
\item We distill a massive synthetic training dataset generated by an autoregressive AI model into a long-range forecast model.
\item Distilled model achieves S2S forecast skill comparable to ECMWF IFS ensemble.
\end{keypoints}

\begin{abstract}
    Accurate long-range weather forecasting remains a major challenge for AI models, both because errors accumulate over autoregressive rollouts and because reanalysis training datasets provide limited sampling of the slow modes of variability underpinning predictability. Most AI and all numerical weather prediction models are autoregressive, repeatedly applying short-lead forecasts to reach subseasonal-to-seasonal (S2S) or seasonal lead times, often causing instability and miscalibration. Long-timestep probabilistic models that generate long-range forecasts in a single step offer an alternative, but training on the 40-year reanalysis record leads to overfitting, suggesting orders of magnitude more data are required. We introduce \textit{long-range distillation}, whereby a long-timestep probabilistic ``student'' model is trained to forecast directly at long-range using a huge synthetic dataset generated by an autoregressive ``teacher'' model. Using the Deep Learning Earth System Model (DLESyM) as teacher, we generate over 10,000 years of simulated climate to train student models across a range of timescales. In perfect-model experiments, the distilled models outperform climatology and approach their teacher's skill while replacing hundreds of autoregressive steps with a single timestep. In the real world, both a diffusion model student fine-tuned on ERA5 and a linear inverse model student with mixed DLESyM-ERA5 training achieve S2S forecast skill competitive with the ECMWF IFS ensemble, with synthetic pre-training providing improvements beyond ERA5-only training. This demonstrates the benefit of synthetic pre-training for both deep learning and classical data-driven methods, and represents the first demonstration that real-world skill can be improved by training on AI-generated synthetic data in volumes far exceeding the observational record.
\end{abstract}

\section*{Plain Language Summary}

Predicting weather weeks to months ahead is extremely challenging. Most current AI weather models extend short forecasts by repeatedly stepping forward in time, causing errors to accumulate and long-range predictions to become unstable. In addition, observational records span only a few decades, limiting the ability to learn slow climate processes that control long-range predictability.

We introduce long-range distillation, an approach that trains an AI model to make long-range forecasts in a single step. We first use an existing AI weather model to generate over 10,000 years of synthetic climate data, enabling the distilled model to learn patterns not well represented in the limited observational record.

The resulting model produces stable long-range forecasts using one step instead of hundreds. In controlled experiments, it nearly matches the performance of the model that generated the synthetic data, and when adapted to real-world conditions, it achieves skill comparable to the ECMWF subseasonal-to-seasonal system. Forecast skill improves with increasing synthetic training data, demonstrating that AI-generated climate simulations can be used to scale long-range prediction skill.

\section{Introduction}

Data-driven approaches to weather forecasting have explored two broad paradigms for mapping an initial atmospheric state to a forecast: \textit{direct} models, which predict the atmospheric state at a target lead time in a single forward pass \cite{rasp2021data,sonderby2020metnet,clare2021combining}, and \textit{autoregressive} models, which learn a short-timestep (typically a few hours) dynamical surrogate and produce long-range forecasts by repeated iterative application \cite{dueben2018challenges,weyn2019can}. While early studies demonstrated the viability of both approaches, in recent years the field has largely adopted autoregressive models, which now match or exceed the forecast skill of state-of-the-art dynamical models for medium-range weather forecasting \cite{pathak2022fourcastnet,keisler2022forecasting,bi2023accurate,chen2023fengwu,chen2023fuxi,lam2023learning,price2025probabilistic,alet2025skillful}. This has been enabled by the availability of reanalysis data from ERA5 \cite{hersbach2020era5}, which provides a large and diverse library of short-range atmospheric trajectories, allowing the training of expressive neural network architectures which generalize well to unseen weather states. Forecasting at long lead times, including beyond those optimized during training, is achieved by repeatedly calling the trained model for multiple iterative time steps. The repeated model calls lead to an accumulation of model errors when used for longer lead times, though this can be partly alleviated by fine-tuning with a multi-timestep objective \cite{weyn2020improving,pathak2022fourcastnet,lam2023learning,nguyen2024scaling,price2025probabilistic,bonev2025fourcastnet} or by coupling to an ocean model and fine-tuning the full coupled system \cite{duncan2025samudrace}.

Despite the impressive successes of autoregressive models, they suffer a number of key limitations when applied to long-range weather forecasting, for example at subseasonal-to-seasonal (S2S) and seasonal lead times. Autoregressive models must be iteratively rolled out over many steps for long-range forecasting, typically far beyond the horizon over which they were trained, often leading to rollout instability when errors accumulate. At S2S and seasonal lead times, predictability from initial conditions diminishes and the skill of deterministic models saturates to climatology, necessitating a probabilistic approach. To produce probabilistic forecasts at long lead times, autoregressive models can be combined with ensembling methods from numerical weather prediction such as perturbed or lagged initial condition ensembles \cite{brenowitz2025practical}, or AI-specific approaches like model checkpoint ensembles \cite{weyn2021sub,mahesh2025henspart1,mahesh2025henspart2}. Recent studies show that ensembles generated using autoregressive models can achieve similar skill at S2S time-scales to the ECMWF IFS ensemble forecasting system \cite{weyn2024ensemble,chen2024machine,lang2024aifs}. However, such ensembling methods have thus far yielded only modest S2S skill gains compared to ECMWF IFS ensemble and calibrating the spread of the ensemble at long lead times is an involved and delicate procedure. Alternatively, probabilistic predictions can be achieved using probabilistic deep learning models, for example diffusion models \cite{mardani2025residual,price2025probabilistic,brenowitz2025climate} or by training an ensemble of perturbed models with probabilistic scoring objectives \cite{lang2024aifs,alet2025skillful,bonev2025fourcastnet}. However, past studies applying probabilistic models still work mostly within an autoregressive framework and are thus still limited in their long-range forecast skill by error accumulation and are hard to calibrate at long time horizons. Fundamentally, there is a mismatch between the autoregressive training objective, learning a short lead timestepper, and the intended inference task in long-range weather forecasting: a well calibrated probabilistic model of the slow modes of climate variability that underpin S2S and seasonal predictability. 

While autoregressive modeling has become common practice in AI weather forecasting, direct models using a single, large model timestep is another viable avenue explored in some early studies \cite{sonderby2020metnet,rasp2021data}, including for probabilistic forecasts \cite{clare2021combining}. Training direct, single timestep probabilistic models to forecast well calibrated ensembles at long lead times appears a promising approach since it removes the risk of autoregressive rollout instability and calibrating the predictions of a single-step model would likely be less challenging than for an ensemble of autoregressive rollouts. Further, direct modeling approaches offer a capability with no analogue in numerical weather prediction: the ability to learn statistical relationships between initial conditions and the slow modes of climate variability that underpin S2S and seasonal predictability, bypassing the intervening dynamical evolution entirely. This represents an underexplored but promising region of the design space for long-range AI weather forecasting.

Despite their appealing properties for long-range forecasting, training single timestep models for S2S and seasonal time-scales on ERA5 reanalysis data presents a data sparsity problem not encountered for short- and medium-range forecasting. At long lead times, the number of independent training examples within the ERA5 record diminishes by orders of magnitude due to the significant autocorrelation at long time-scales. ERA5 contains relatively few independent realizations of S2S variability, and even fewer of seasonal variability. This poses a challenge for training expressive neural network architectures without overfitting to the limited range of dynamics sampled in the ERA5 record.

In this study, we propose a novel approach to probabilistic long-range weather modeling: {\em long-range distillation}. Our approach is inspired by the fact that short timestep autoregressive models are good generators of realistic atmospheric variability, but that controlling an ensemble of such models for well calibrated long-range forecasting is challenging. We propose to use a short timestep autoregressive model as a `teacher' model to enable the training of a probabilistic long timestep `student' model through synthetic training data. Autoregressive models are fast and cheap to run and, provided they can be rolled out stably over long time-horizons, can thus simulate orders of magnitude more years of atmospheric variability than is available in ERA5. By generating a large volume of simulated weather states using the autoregressive teacher model, we aim to construct a sufficiently large and diverse training dataset to train a probabilistic model to produce well calibrated long-range forecasts in a single model timestep. The trained student model can then be thought of as a distilled representation of the autoregressive teacher simulations. By using a single model timestep, the distilled model would provide a controllable interface for long-range modeling which sidesteps the need for a large number of autoregressive time steps and for extensive tuning of initial condition or model weight perturbations for well calibrated ensemble forecasts. 

The performance of large language models has been shown to scale as both the training dataset size and model capacity are increased by orders of magnitude, with optimal results obtained when both are scaled simultaneously \cite{kaplan2020scaling}. AI weather models have thus far explored scaling model capacity while keeping training dataset size fixed, restricted by the limited duration of the ERA5 record. By using computationally cheap autoregressive models to generate large volumes of synthetic weather data, we aim to unlock the dataset size scaling axis for our distilled model training. Our departure away from autoregressive modeling and toward compressing a large volume of climate data into a conditional generative model is also inspired by recent efforts to develop a generative foundation model of the global atmosphere \cite{brenowitz2025climate}. 

Training distilled models for long-range prediction requires an autoregressive teacher model capable of generating a large volume of synthetic training data with physically plausible S2S and seasonal variability. Here, we use the Deep Learning Earth System Model (DLESyM) \cite{cresswell2025deep} as the autoregressive teacher model for our distilled, long timestep student models. DLESyM is a coupled ocean-atmosphere autoregressive model trained on reanalysis data, with the ocean and atmosphere models trained separately before being coupled at inference. Despite only being optimized over short time horizons of a few days, DLESyM remains stable when rolled out for 100's-1000's of years and produces emergent modes of coupled ocean-atmosphere variability such as the El-Nino Southern Oscillation (ENSO), with realistic climatological statistics for a wide range of processes such as tropical cyclogenesis, the Indian Summer Monsoon, and mid-latitude blocking events \cite{cresswell2025deep}. While we use DLESyM here, we emphasize that our long-range distillation framework could be applied in future to any autoregressive model capable of realistic long-range rollouts, for example SamudrACE \cite{duncan2025samudrace,watt2025ace2,dheeshjith2025samudra} or NeuralGCM \cite{kochkov2024neural}. Another advantage of DLESyM as the autoregressive teacher is that it is trained exclusively on reanalysis data, ensuring that our distilled models learn dynamics grounded in observations rather than inheriting the inherent biases of physics-based climate models, which have complementary advantages for forced multi-decadal prediction tasks. 

While DLESyM is a promising autoregressive teacher model for long-range distillation due to its long-term stability and coupled ocean–atmosphere variability, it has limitations relevant to this study. First, DLESyM simulates a limited set of state variables, just 8 atmospheric variables and a single ocean variable, compared with roughly 100 atmospheric variables in state-of-the-art medium-range AI weather models \cite{pathak2022fourcastnet,keisler2022forecasting,bi2023accurate,chen2023fengwu,chen2023fuxi,lam2023learning,price2025probabilistic,alet2025skillful}, where the same variable at different pressure levels is counted separately. While this parsimonious state vector promotes long-term stability, it potentially restricts the range of dynamics the model can represent and could limit forecast skill relative to larger models. Second, recent studies have shown that autoregressive AI models struggle to capture the rapid initial growth of small-amplitude perturbations in the initial conditions \cite{selz2023can}. This has implications for ensemble forecasting using perturbed initial conditions, as considered in Section \ref{methods:model_exp}, because it suggests that perturbation strengths may need to be artificially increased beyond the true uncertainty in initial conditions to compensate for suppressed error growth. \citeA{selz2023can} found that the Pangu model \cite{bi2023accurate}, while failing to reproduce the initial fast growth of small perturbations compared with a physics-based model ground truth, largely captured the true error growth for larger initial condition perturbation sizes comparable to the uncertainty of real-world analysis. In contrast, our experiments (Section \ref{methods:model_exp}) suggest that DLESyM exhibits suppressed error growth even for large initial perturbations, likely due to its limited state vector, a point warranting further investigation beyond the scope of this study. The key implication for this work is that the predictability of the DLESyM-simulated atmosphere differs from the real world. This should be kept in mind when interpreting the results of the perfect-model experiment in Section \ref{methods:model_exp}, though we expect the difference to be most pronounced at short- to medium-range lead times, where sensitivity to initial conditions is strongest, and less so at S2S or seasonal scales.

In this study, we explore the feasibility of long-range distillation for probabilistic forecasting across a range of time-scales through controlled perfect-model experiments, before applying our framework to real-world S2S forecasting. Section 2 outlines our proposed long-range distillation method. Section 3 describes the setup and results of our perfect-model experiment. Section 4 describes how we fine-tune and apply our distilled models for real-world forecasting and benchmarks our method against the ECMWF IFS ensemble forecast system. We draw conclusions and discuss future research directions in Section 5.

\section{Long-Range Distillation}

\subsection{Long-Range Distillation Problem Formulation}\label{methods:distillation_framework}

\begin{figure}[t]
    \centering
    \includegraphics[width = \columnwidth]{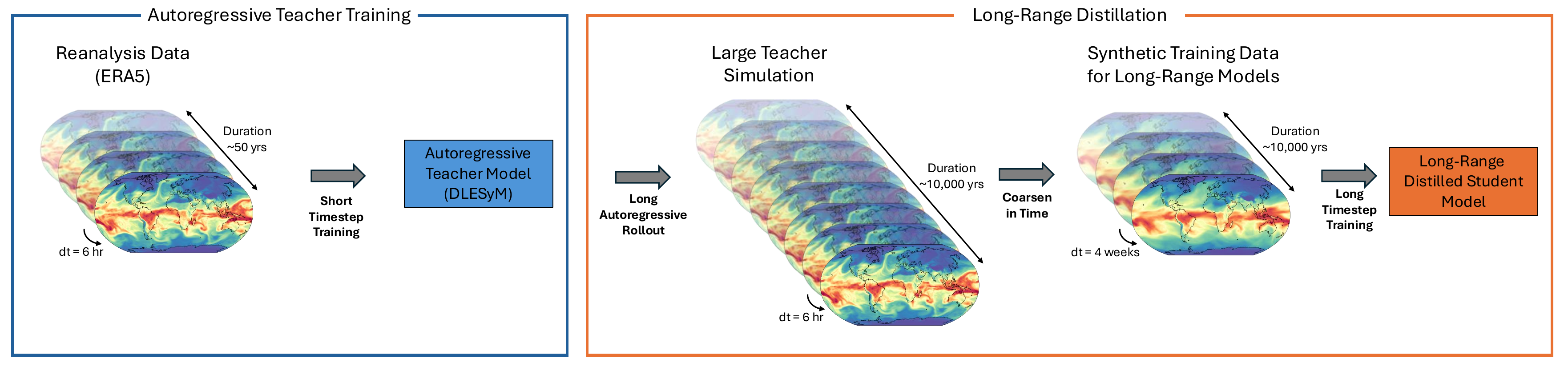}
    \caption{Schematic of long-range distillation approach. First, an autoregressive teacher model is trained for short lead time prediction using reanalysis datasets like ERA5 --- in our case we leverage DLESyM as a pre-trained autoregressive teacher capable of stable long-running simulations. The autoregressive teacher is used to generate a huge simulation with orders of magnitude higher temporal coverage than ERA5. Finally, a long timestep probabilistic model, the "student", is trained on the autoregressive teacher to do long-range prediction in a single model timestep.}
    \label{fig:method_schematic}
\end{figure}

Long-range distillation assumes access to a short time-step autoregressive model trained on reanalysis data capable of stable long rollouts, which here will be DLESyM (Section \ref{methods:dlesym_data_generation}). This autoregressive `teacher' model is used to train a distilled `student' model to do long-range forecasting in a single model timestep (Figure \ref{fig:method_schematic}). 

First, a large simulation is created using the teacher model, generating a diverse dataset of synthetic input-output pairs for forecasting at long time-scales. This dataset contains orders of magnitude more long-range training examples than the original reanalysis dataset on which the autoregressive teacher was trained.

Given a long rollout from the autoregressive teacher, ($x_{1}$, $x_{2}$, ...), we first define our long-range target, $\overline{x}_N$, which is the teacher-simulated state at some large number, $N$, of autoregressive steps forward and averaged over a time window of $M$ autoregressive steps,
\begin{equation}\label{eqn:target_definition}
    \overline{x}_N = \frac{1}{M}\sum_{i=-M/2}^{i=M/2}x_{N+i}.
\end{equation}
Our long-range distilled model then seeks to directly model the conditional probability of this long-range target
\begin{equation}
    p(\overline{x}_N \vert x_1),
\end{equation}
conditioned on the initial state. We thus replace a long autoregressive rollout of potentially hundreds of steps with a single model timestep.

Concretely, at S2S time-scales, given an autoregressive model which takes 6 hr timesteps, we could choose $N=112$ and $M=28$ to target a weekly average forecast at 4-week lead time. In practice, to reduce the data volume of the large autoregressive teacher simulation we save the teacher simulation with degraded temporal resolution, averaging the 6-hourly timesteps to daily frequency. In practice, we then condition our distilled model forecasts on a short history of four daily averages rather than a single day. Although the synthetic training data are generated using a Markov model (i.e., DLESyM), and a history of states should therefore carry no extra information, we include it to compensate for information lost through daily averaging.

\subsection{DLESyM Climate Simulation}\label{methods:dlesym_data_generation}

We use DLESyM \cite{cresswell2025deep} to generate O(10,000 years) of synthetic climate for training our long-range distilled models. To parallelize data generation, we choose to generate an ensemble of simulations initialized on different dates rather than running a single long-running simulation. We initialized DLESyM simulations on 200 dates equally spaced between 2008-01-01 and 2016-12-31, running each simulation for 90 years, yielding 18,000 years of simulated climate. We used the version of DLESyM distributed through the NVIDIA Earth2Studio package (v0.8.0), with each ensemble member using a randomly selected checkpoint from the four checkpoints provided through the NGC catalog \cite{dlesym_ngc} to boost ensemble variability. The Earth2Studio implementation of DLESyM closely follows that outlined in \citeA{cresswell2025deep}, except that the atmospheric model omits outgoing longwave radiation from the set of prognostic variables predicted, with the model checkpoints being trained again from scratch following the procedure outlined in \citeA{cresswell2025deep}. DLESyM is initialized using reanalysis data from ERA5 regridded onto the HEALPix64 grid. The prognostic variables predicted by this version of DLESyM are:
\begin{itemize}
    \item Sea surface temperature (SST)
    \item 2 m air temperature ($T_{2m}$)
    \item Temperature at 850hPa ($T_{850}$)
    \item Total column water vapor (TCWV)
    \item Wind speed at 10m ($w_{10m}$)
    \item Geopotential height at 1000hPa ($z_{1000}$)
    \item Geopotential height at 500hPa ($z_{500}$)
    \item Geopotential height at 250hPa ($z_{250}$)
    \item Difference between geopotential heights at 700hPa and 300hPa ($\tau_{300-700}$)
\end{itemize}

\begin{figure}[t]
    \centering
    \includegraphics[width = 0.8\columnwidth]{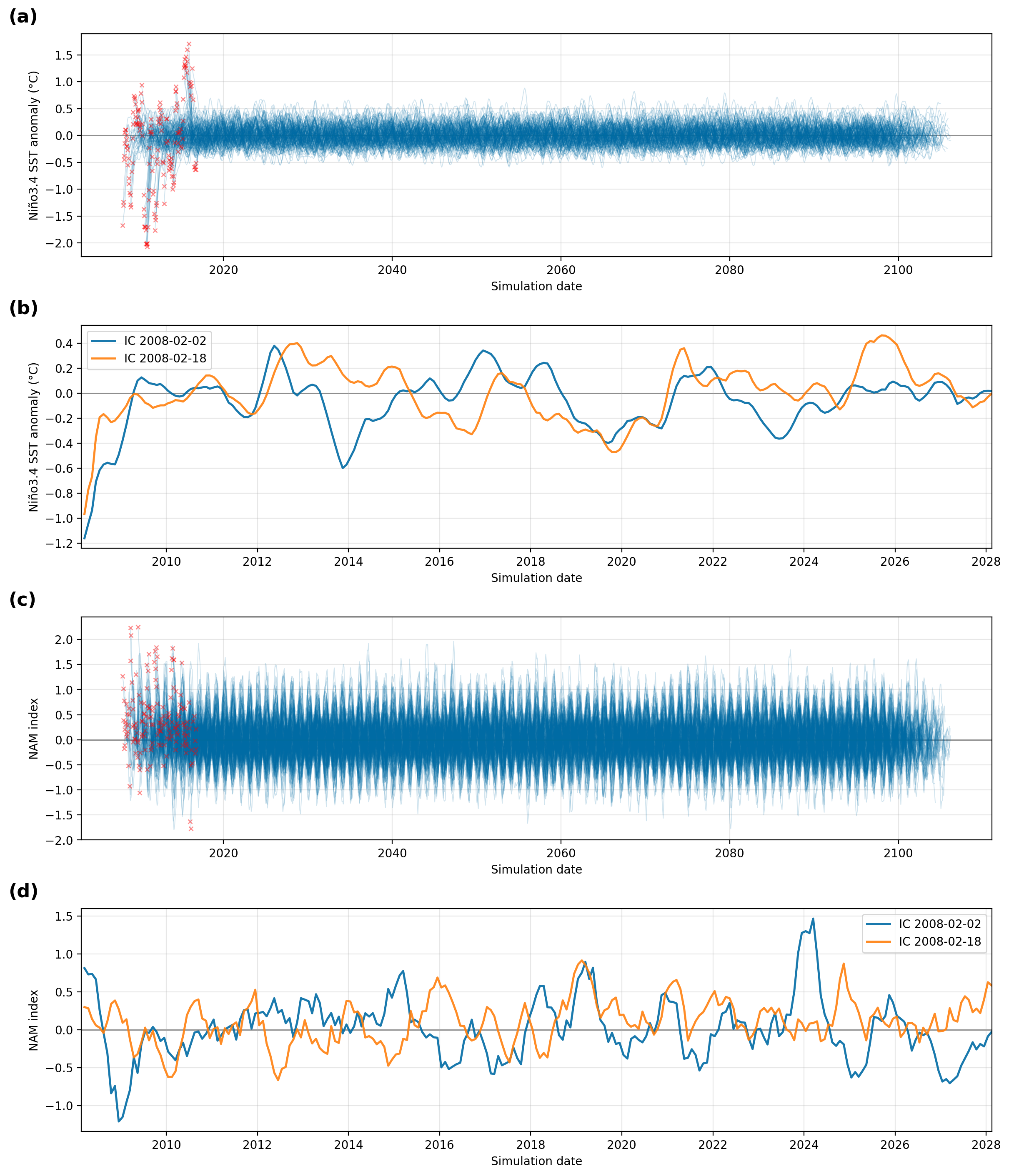}
    \caption{DLESyM ensemble simulation training data. (a) Nino3.4 index calculated for each of the DLESyM ensemble members throughout the 90-year simulation with the corresponding initialization date denoted by a red cross. (b) Zoomed in view of first 20 years of 2 ensemble members initialized 16 days apart showing how after initial spin-up, they quickly diverge and generate independent training data samples despite being initialized from similar initial conditions. (c) Same as (a) but for the Northern Annular Mode (NAM). (d) Same as (b) but for NAM.}
    \label{fig:dlesym_simulations}
\end{figure}

Generating this large ensemble of DLESyM simulations took 4 hours when run in parallel on 96 NVIDIA H100 GPUs, with a throughput of O(1000) simulated years per day per H100 GPU. This emphasizes the substantial speedup offered by AI climate emulators compared to physics-based climate models. We observed that $\sim$14\% of the DLESyM ensemble members went unstable before finishing the 90 year rollout. This is in contrast to the 1,000 year stable rollouts demonstrated by \citeA{cresswell2025deep} and this infrequent instability appears to be specific to the Earth2Studio DLESyM checkpoints as running the checkpoints from \citeA{cresswell2025deep} for the initial conditions that went unstable here did not show signs of instability in the first 100 years of simulation \cite{dlesym_stability_communication}. Regardless, it is easily identified and pruned in post-processing, i.e. we simply omit all ensemble members that went unstable during the 90 year simulation, leaving $\sim$15,000 years of simulated climate for training and validation of our distilled models. Details of the DLESyM ensemble simulation are summarized in \ref{sec:hyperparams}. Despite being initialized from closely-spaced ERA5 initial conditions, each member of the DLESyM ensemble after a brief spin-up period diverges, so that each ensemble member provides independent samples of the DLESyM model's climate (Figure \ref{fig:dlesym_simulations}). This is readily apparent both for tropical (panels a \& b) and extra-tropical (panels c \& d) variability.

\subsection{Distilled Student Model Architecture: HEALPix Conditional Diffusion}\label{methods:nn_architecture}

We implement our distilled student model as a conditional diffusion model, enabling probabilistic forecasting. As input conditioning, $c$, we provide a short history of daily average weather states for the 4 preceding days. We then seek to model the conditional distribution, $p(x\vert c)$, of future global weather states, $x$, at some long-range target lead time. All data are represented on the HEALPix64 grid as it provides a satisfying discretization of the sphere and is consistent with the DLESyM autoregressive teacher model. The neural network architecture we use to perform global conditional diffusion on the HEALPix64 grid follows that used in the recently-developed global foundation model, \textit{cBottle} \cite{brenowitz2025climate}. The reader is referred to \citeA{brenowitz2025climate} for implementation details. Briefly, this architecture adapts the UNet backbone used widely in diffusion models \cite{song2020score,karras2022elucidating} to the HEALPix grid by using a HEALPix-specific padding scheme proposed by \citeA{karlbauer2024advancing}, by adding a learned spatial embedding to learn non-stationary spatial structures, and by adding a periodic temporal embedding to represent the day of year and time of day \cite{brenowitz2025climate}. Since in this study all forecast and conditioning states are averaged over at least one day, we drop the time of day embedding used in \textit{cBottle} but retain the day of year embedding. We train using a log uniform $\sigma$ distribution and for sampling we use $\sigma_{min}=0.002$, $\sigma_{max}=200$, and 18 sampler steps. Note these settings differ from those in the original EDM paper \cite{karras2022elucidating} and were chosen following the rationale laid out in \citeA{brenowitz2025climate} to ensure that the noise range fully covers all scales in our data and to ensure the training $\sigma$ distribution matches the sampler scheme used at inference. All network and sampler hyperparameters are listed in \ref{sec:hyperparams}.

\subsection{Calibrating Distilled Model Forecast with Classifier-Free Guidance}\label{methods:classifier_free_guidance}

Calibrating forecasts requires a mechanism to control the ensemble spread of the forecast. For an initial condition ensemble of autoregressive models, this is typically achieved by varying the strength of the initial condition perturbations. To control the ensemble spread from our long timestep conditional diffusion models we here exploit classifier-free guidance \cite{ho2022classifier}. Classifier-free guidance provides a way to achieve a trade-off between the diversity of generated samples and the strength of the conditioning.

Concretely, the conditional score, $\nabla_x\text{log}(p(x(t)\vert c))$, used during sampling, is replaced with a linear combination of the conditional and unconditional scores,
\begin{equation}
     \nabla_x\text{log}\left(p(x(t))\right) + w\left[\nabla_x\text{log}(p(x(t)\vert c)) 
- \nabla_x\text{log}(p(x(t)))\right],
\end{equation}
where $w$ is the classifier-free guidance weight that controls the strength of the guidance. This is equivalent to sampling from the probability density function $p(x(t)|c)^wp(x(t))^{1-w}$. When $w=0$, the model generates samples unconditionally. When $w=1$, the model generates samples conditionally with no guidance. When $w>1$ the model is guided strongly towards the conditioning, and when $0<w<1$ the conditioning strength is reduced. In practice, we jointly learn the conditional and unconditional models with the same network by randomly setting the input conditioning tensor to zero during training for 10\% of the samples. While classifier-free guidance is common practice in computer vision, this is to our knowledge the first study to propose using it to control ensemble spread in weather forecasting. Unlike computer vision, where classifier guidance is used to increase fidelity to conditioning by choosing $w > 1$, we typically use the guidance to increase ensemble spread by choosing $w<1$  (see Figure \ref{fig:cfg_spread}a).

\section{Perfect-Model Experiment}

\subsection{Setup}\label{methods:model_exp}

In Section \ref{real_world_methods} we turn to the problem of real-world forecasting and account for the domain shift needed to apply the DLESyM-trained student model to real-world ERA5 data. First, we evaluate the ability of our long-range distilled models to produce skillful ensemble forecasts across a range of lead times in the controlled setting of a perfect-model experiment. All evaluation is done using a withheld simulation from the autoregressive teacher model (DLESyM). We assess the extent to which both the distilled long timestep models and the autoregressive DLESyM teacher model can forecast an unseen DLESyM simulation given imperfect knowledge of initial conditions. 

Concretely, we apply the following experiment protocol:
\begin{enumerate}
    \item Perform a single, multi-year deterministic simulation with DLESyM  referred to hereafter as the Nature Run. This simulation serves as the ground truth reference and is assumed to represent the unknown, true state of the atmosphere.
    \item Select a series of starting-point dates from the Nature Run. Apply a different set of initial condition perturbations to each of these starting points  to create the suite of  imperfect initial conditions for each ensemble forecast.
    \item For the distilled long-timestep student model, construct ensemble forecasts using the conditional diffusion model applied to the imperfect initial conditions from 2. The spread of the ensemble forecast can be controlled using classifier-free guidance (Section \ref{methods:classifier_free_guidance}).
    \item For the DLESyM autoregressive teacher model, construct either deterministic or ensemble forecasts from the same imperfect initial conditions from 2 as used by the distilled model. To construct an ensemble forecast from DLESyM, we use initial condition perturbations (perturbing around the already imperfect initial conditions from 2) to generate ensemble spread. The spread of the ensemble forecast can be controlled by varying the strength of the extra initial condition perturbations applied on top of those from 2.
    \item Score all forecasts against the unseen Nature Run from 1.
\end{enumerate}

The aim of this experiment is to test the extent to which our distilled models produce a well-calibrated ensemble forecast, and we compare its performance against an ensemble forecast made using the DLESyM autoregressive teacher model through initial condition perturbations for reference. The size of the initial condition perturbation made to the Nature Run when constructing the imperfect initial conditions is an important design choice of our experiment. Since DLESyM is a deterministic autoregressive model, in the limit where we provide a perfect initial condition it will provide a perfect forecast. Thus we seek to provide a large enough perturbation to ensure the experiment reflects the practical real-world forecasting ability of DLESyM given imperfect knowledge of initial conditions. To achieve this, we tune the size of the initial condition perturbation until the RMSE in $Z_{500}$ at 7-day lead time of a DLESyM ensemble member matches that of IFS HRES (521 $m^2$/$s^2$) \cite{rasp2024weatherbench} (Figure \ref{fig:ic_error_growth}). This corresponds to an initial condition perturbation size of 65 $m^2$/$s^2$ for $Z_{500}$. We acknowledge this perturbation is larger than the likely real-world analysis error, IFS HRES has 22 $m^2$/$s^2$ RMSE at 6-hour lead time \cite{rasp2024weatherbench}, which we interpret as necessary to compensate for the known slower error growth of autoregressive AI weather models compared to dynamical models \cite{selz2023can}. We perturb all atmospheric variables by the same fraction of their global standard deviation which is the single parameter we tune to match day-7 IFS HRES RMSE. For SST we provide a perturbation of 0.1 K. The initial condition perturbations are sampled as Gaussian white noise with spatiotemporal correlation structure given by a Matern covariance function with de-correlation length-scale 500 km and time-scale 48 hours, using the implementation in Earth2Studio \cite{e2s_pert_code}. This noise structure, which follows that used by \citeA{leutbecher2008ensemble}, encourages the perturbed fields to retain physically realistic spatiotemporal structure, while being significantly simpler to implement than dynamical methods such as bred vectors \cite{toth1993ensemble} given that perturbed initial condition ensembling is not a primary focus of our study.

\begin{figure}[t]
    \centering
    \includegraphics[width = \columnwidth]{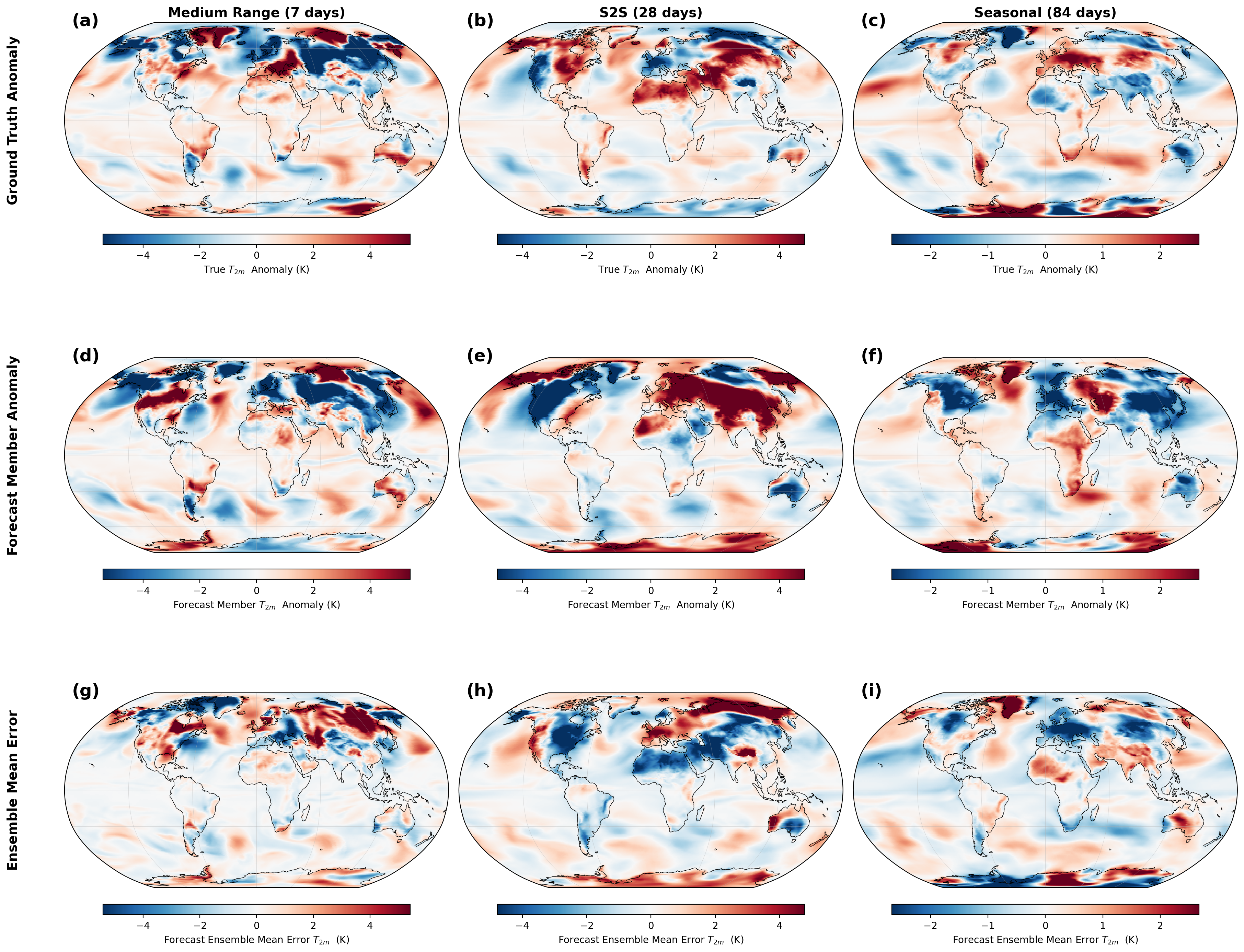}
    \caption{Example distilled student model 2 m air temperature forecasts across a range of lead times initialized on 2017-01-01. Ground truth forecast targets from the withheld DLESyM evaluation simulation or (a) medium-range, (b) S2S, and (c) seasonal lead times. (d-f) Single forecast member predictions for each lead time. Both forecast targets and ground truth targets are visualized as anomalies from climatology (a-f). (g-i) Ensemble mean forecast error for a 32-member ensemble forecast for each lead time. Medium-range forecasts (a, d, g) are daily means at 7-day lead time, S2S forecasts (b, e, h) are weekly means at 4-week lead time, and seasonal forecasts (c, f, i) are 4-week means at 12-week lead time.}
    \label{fig:forecast_viz}
\end{figure}

All validation is done on a withheld DLESyM simulation not seen during training of the distilled model. In addition to deterministic metrics like ensemble RMSE, we score all models using continuous ranked probability score (CRPS) \cite{hersbach2000decomposition}. We test distilled long timestep models at three distinct lead times (Table \ref{tab:lead_times}), representing three distinct predictability regimes (Figure \ref{fig:forecast_viz}):

\begin{figure}[t]
    \centering
    \includegraphics[width = \columnwidth]{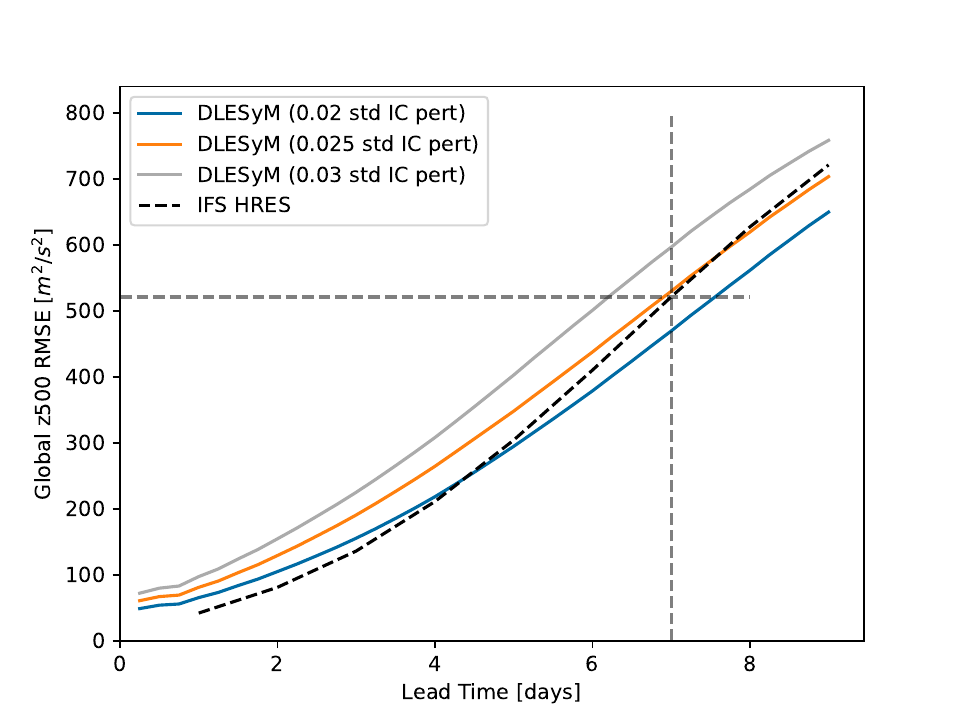}
    \caption{Error growth in DLESyM autoregressive ensemble members for varying initial condition perturbation strengths. Global average $Z_{500}$ RMSE of DLESyM ensemble members with respect to a withheld ensemble member for varying initial condition perturbation sizes (colored lines) and the real-world global average $Z_{500}$ RMSE of IFS for reference (black dashed line). The initial condition condition perturbation strengths in the legend are provided in normalized units since we perturb all fields using a multiple of each field's global standard deviation. For $Z_{500}$, the standard deviation is 2605 $m^2/s^2$.}
    \label{fig:ic_error_growth}
\end{figure}

\begin{itemize}
    \item \textbf{Medium Range:} We forecast a daily average at 7-day lead time ($N=28$, $M=4$). For evaluation, we use over 400 initialization dates spaced 2 days apart from 2017-01-01 to 2019-03-10 in simulation years.
    \item \textbf{S2S:} We forecast a weekly average at 4-week lead time ($N=112$, $M=28$). For evaluation, we use over 400 initialization dates spaced 4 days apart from 2017-01-01 to 2021-05-16 in simulation years.
    \item \textbf{Seasonal:} We forecast a monthly average at 12-week lead time ($N=336$, $M=112$). For evaluation, we use over 400 initialization dates spaced 8 days apart from 2017-01-01 to 2025-09-28 in simulation years.
\end{itemize}

\begin{table}\label{tab:lead_times}
\caption{Inputs and outputs for the different distilled student models. $N$ is the lead time in number of 6-hour autoregressive teacher steps and $M$ is the size of the target averaging window in 6-hour teacher steps (Equation \ref{eqn:target_definition}). Inputs beyond Day 0 were added to compensate for the information lost by averaging the 6-hourly DLESyM teacher model data to daily resolution during data generation.}
\centering
\begin{tabular}{l c c c c}
\hline
 Lead Time  & Inputs & Outputs & N  & M  \\
\hline
  Medium-Range  & Days 0, -1, -2, -3  & Day 7 & 28 & 4   \\
  S2S  & Days 0, -1, -2, -3  & Average of Days 25-31 & 112 & 28   \\
  Seasonal  & Days 0, -1, -2, -3  & Average of Days 70-98 & 336 & 112   \\
\hline
\multicolumn{3}{l}{}
\end{tabular}
\end{table}

We split the 15,000 years of DLESyM simulation into training (75\%) and validation (25\%) datasets, splitting by ensemble member in our large ensemble to ensure separation between training and validation datasets. After applying our train-validation split, we are left with $\sim$11,000 years of DLESyM simulations for training. For each lead time, we train a separate distilled model. See \ref{sec:hyperparams} for more details about the training.

In Figure \ref{fig:forecast_viz} we show an example forecast (single ensemble member) at each lead time for 2 m air temperature. As the lead time increases, the targets and forecasts become increasingly smooth due to the increasing temporal averaging window, $M$. At medium-range (panels a, d, \& g), there is agreement between the patterns of variability in both the target and forecast, consistent with the relatively high predictability of the atmosphere from initial conditions at this lead time. At S2S (panels b, e, \& h) and seasonal (panels c, f, \& i), the discrepancies between the ensemble member forecast and targets become larger, emphasizing the need for ensemble forecasting at these lead times. The forecasts and targets show the closest agreement over oceans, with predictability likely attributable to the relatively slow evolution of sea surface temperature since ocean dynamics evolves slower than the atmosphere and sea surface temperature and surface air temperature are closely connected. Interestingly, the seasonal forecast (panel f) appears to show a consistent ENSO phase to the target, with elevated surface temperature in the tropical Pacific, suggesting the distilled model is learning to leverage ENSO predictability for seasonal forecasting.

All forecasts are compared against two climatological forecast baselines calculated from a withheld 20 years of the DLESyM simulation. First, we refer to a \emph{deterministic} climatology by retrieving the long-range forecast targets, with the requisite temporal averaging for the given $M$, for each of the 20 years and taking the mean. Second, we refer to a \emph{probabilistic} climatology by using all forecast targets from the 20 year reference period together as an ensemble forecast.

\subsection{Results}

\subsubsection{Distilled Model Skill Improves with Increasing Synthetic Training Data Beyond the ERA5 Record}\label{results:model_exp:scaling}



\begin{figure}[t]
    \centering
    \includegraphics[width = 0.8\columnwidth]{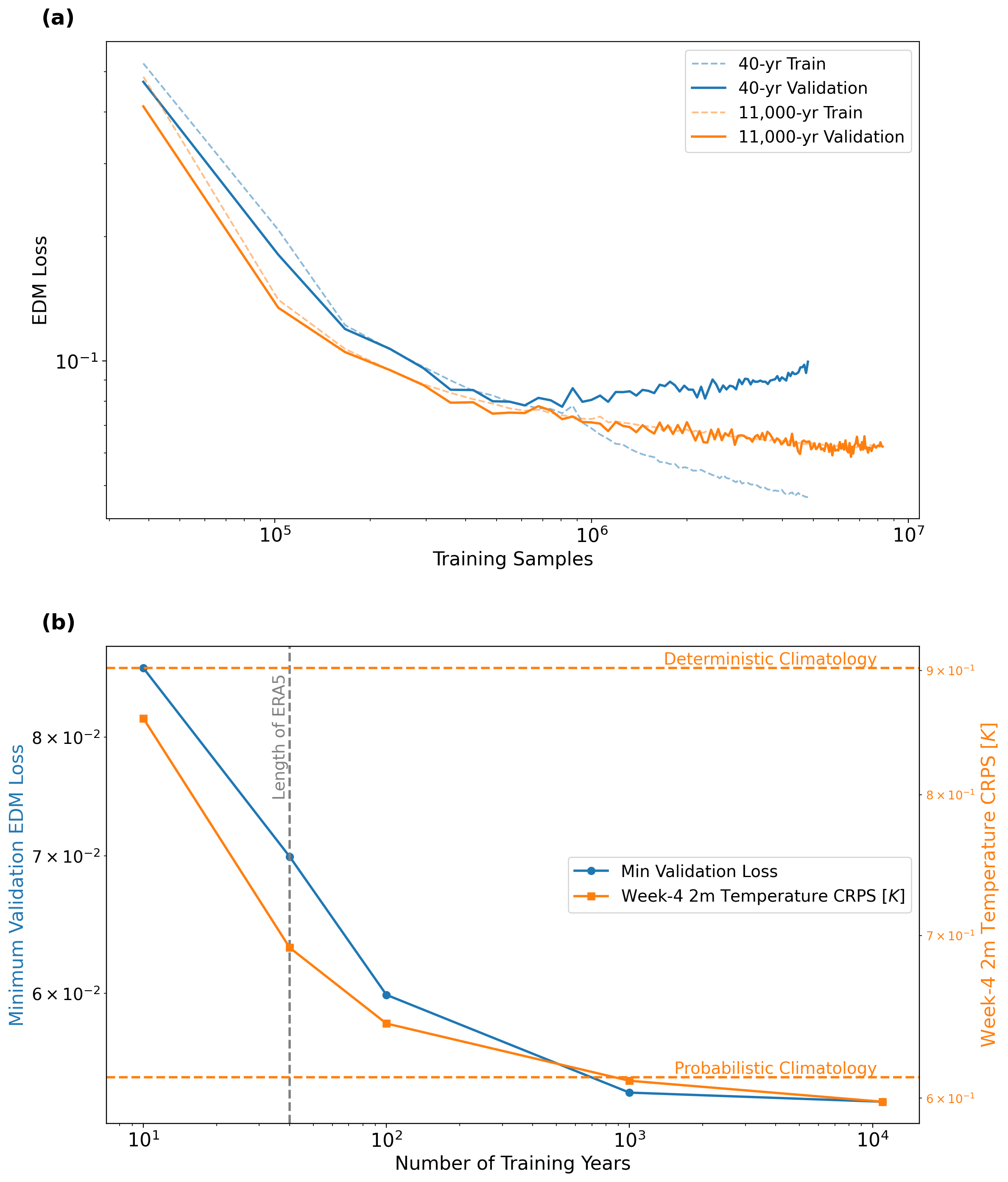}
    \caption{Scaling of distilled student model S2S forecast skill with synthetic training dataset size. (a) Elucidated diffusion model (EDM) loss \cite{karras2022elucidating} throughout training for a 40 year training dataset (blue) and the full 11,000 year dataset (orange). Dashed lines indicate training loss and solid the validation loss. Learning curves have been smoothed for plotting purposes using bin averaging. (b) Distilled model week 4 2 m air temperature global mean CRPS with increasing number of training years (orange) and the corresponding minimum validation EDM loss achieved during training (blue). Dashed lines indicate the length of the ERA5 record and the CRPS skill of climatology for reference. Note all axes are logarithmic.}
    \label{fig:training_scaling_w_years}
\end{figure}

Our long-range distillation framework is based on the hypothesis that an increasing volume of synthetic training data from a long-duration autoregressive teacher simulation improves student skill, even when the teacher simulation duration far exceeds the reanalysis dataset used for training the teacher. For this to be true, the autoregressive teacher simulation must generate a diverse range of climate states beyond those seen during training, sampling internal variability not captured by the short reanalysis record. Analyses of long DLESyM rollouts suggest that it can indeed generate new samples of internal climate variability \cite{cresswell2025deep}, but here we test whether this translates into improved performance of our distilled models with increasing amounts of synthetic training data.

We test this hypothesis by evaluating distilled S2S forecast models trained over a range of autoregressive teacher simulation lengths (Figure \ref{fig:training_scaling_w_years}). A distilled model trained on 40 years of DLESyM simulation, mimicking the availability of ERA5, shows clear signs of overfitting once trained beyond $\sim$1M samples, corresponding to $\sim$15,000 training steps (Figure \ref{fig:training_scaling_w_years}a). By contrast, the model trained on the full 11,000 year DLESyM simulation shows no signs of overfitting, with the training and validation losses matching closely throughout. The reduced overfitting can also be seen in the steadily improving minimum validation loss as the number of training years is increased by orders of magnitude (Figure \ref{fig:training_scaling_w_years}b). Crucially, the improved EDM denoising losses also translate to improvements in S2S forecast skill. Training on 11,000 years of synthetic data yields a 14\% reduction in 2 m temperature CRPS at a 4-week lead time, compared to training on 40 years. Note, the scaling of the forecast skill with dataset size appears to be weaker than the power law scaling curves observed for natural language \cite{kaplan2020scaling}, with our CRPS values not following a power law with dataset size. This is either because beyond a certain point, the simulation becomes large enough to provide comprehensive sampling of the internal variability of the DLESyM model climate, or because the distilled student model capacity becomes the bottleneck as we used a fixed architecture for all dataset sizes whereas \citeA{kaplan2020scaling} scaled both model capacity and dataset size jointly. Nonetheless, the improvements of week-4 CRPS with training data size far beyond the size of the original ERA5 training dataset confirms our hypothesis that generating a large synthetic corpus of training data using autoregressive models enables more skillful long-range models and is the first demonstration of scaling forecast performance with an increasing volume of synthetic training data. 

\begin{figure}[t]
    \centering
    \includegraphics[width = 0.8\columnwidth]{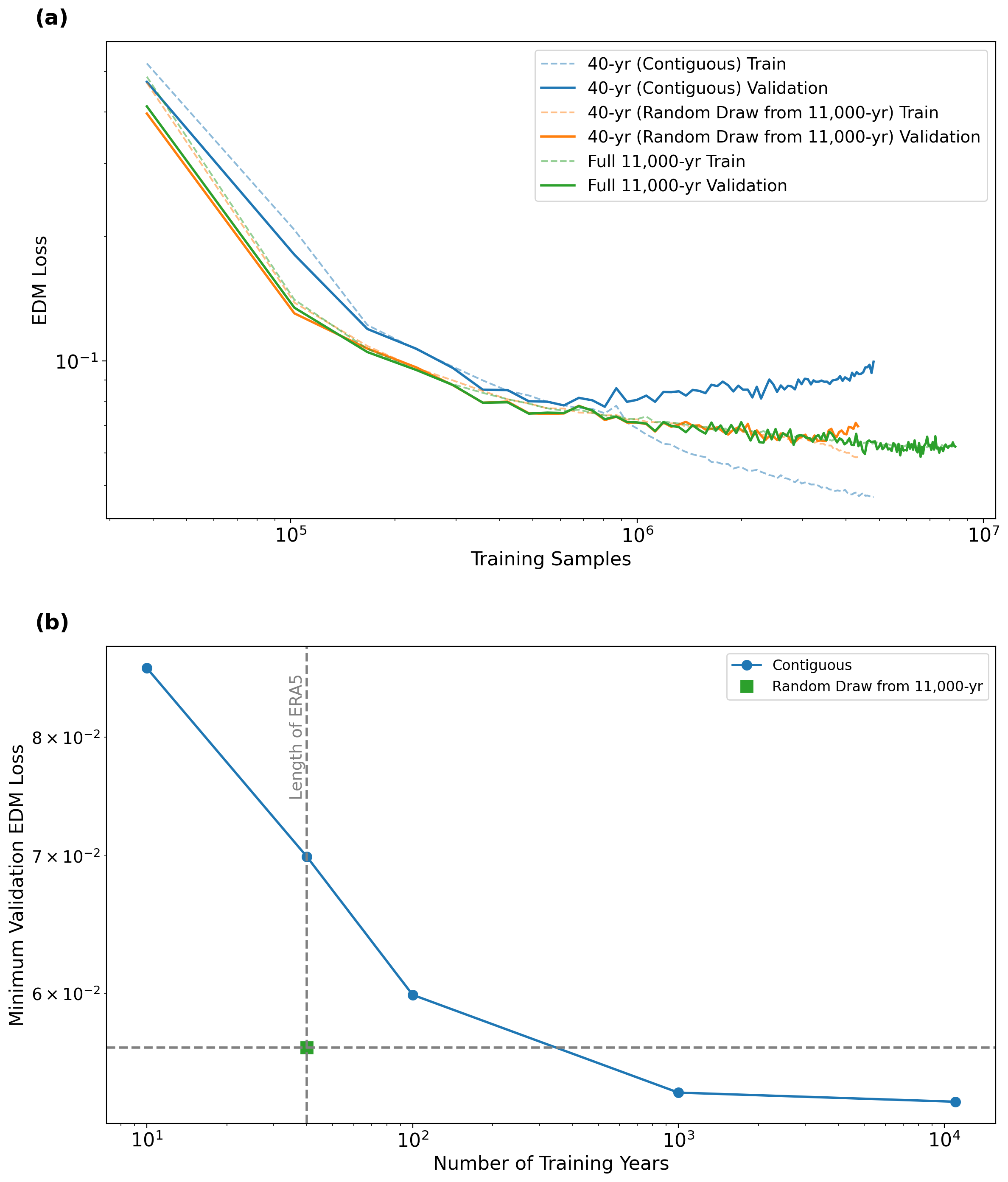}
    \caption{Separating effect of overfitting on small dataset and added sample diversity from large synthetic dataset. (a) EDM loss throughout training for 40 years sampled in a contiguous block to mimic reanalysis record (blue), 40 years sampled randomly from throughout the 11,000 years of simulation output (orange), and for the full 11,000 year dataset (green). Dashed lines indicate training loss and solid the validation loss. Learning curves have been smoothed for plotting purposes using bin averaging. (b) Minimum validation EDM loss achieved during training by training dataset size for contiguously (blue dots) and randomly (from 11,000 years, green squares) sampled training datasets}
    \label{fig:random_draw_ablation}
\end{figure}

The improvement of the student model with increasing number of years of synthetic training data could be driven by two confounding effects: (a) increased regularization from a larger sample count and (b) increased sample diversity, whereby the teacher generates genuinely new, independent samples of the DLESyM climate not captured in a 40-year record. To separate these effects, we train a distilled student model on 40 years of data, matching the reanalysis record in sample count, but drawing samples randomly from throughout the full 11,000-year simulation rather than from a single contiguous block (Figure \ref{fig:random_draw_ablation}). While this randomly-drawn model eventually overfits, it remains regularized for significantly longer than when trained on contiguous years (panel a). The student model trained on 40 years of randomly drawn data even outperforms training on 100 years of contiguous training data despite the lower sample count, lagging only the 1,000-year and 11,000-year models (panel b). This underlines that the benefit of the large synthetic training corpus stems largely from increased sample diversity rather than merely a higher sample count, suggesting the long DLESyM rollouts sample genuinely new states not captured in a 40-year record.

Hereafter, we use the shorthand \emph{DLESyM10K} to refer to the distilled student model trained on the full O(10,000 year) DLESyM simulation and all distilled student model results use the model trained on the full dataset.

\subsubsection{Classifier-Free Guidance Provides a Simple Recipe for Calibration of Long Timestep Diffusion Forecasts}\label{results:model_exp:cfg}



A well calibrated forecast should have a spread-skill ratio near unity. One key motivation for our long-range distillation approach is that forecasting using a single model timestep should make it relatively easy to control the ensemble spread, and hence calibrate the forecasts, compared to controlling ensemble spread over a long autoregressive rollout. Here, we confirm our working hypothesis that classifier-free guidance (Section \ref{methods:classifier_free_guidance}) provides a simple recipe for ensemble calibration.

\begin{figure}[t]
    \centering
    \includegraphics[width = 0.5\columnwidth]{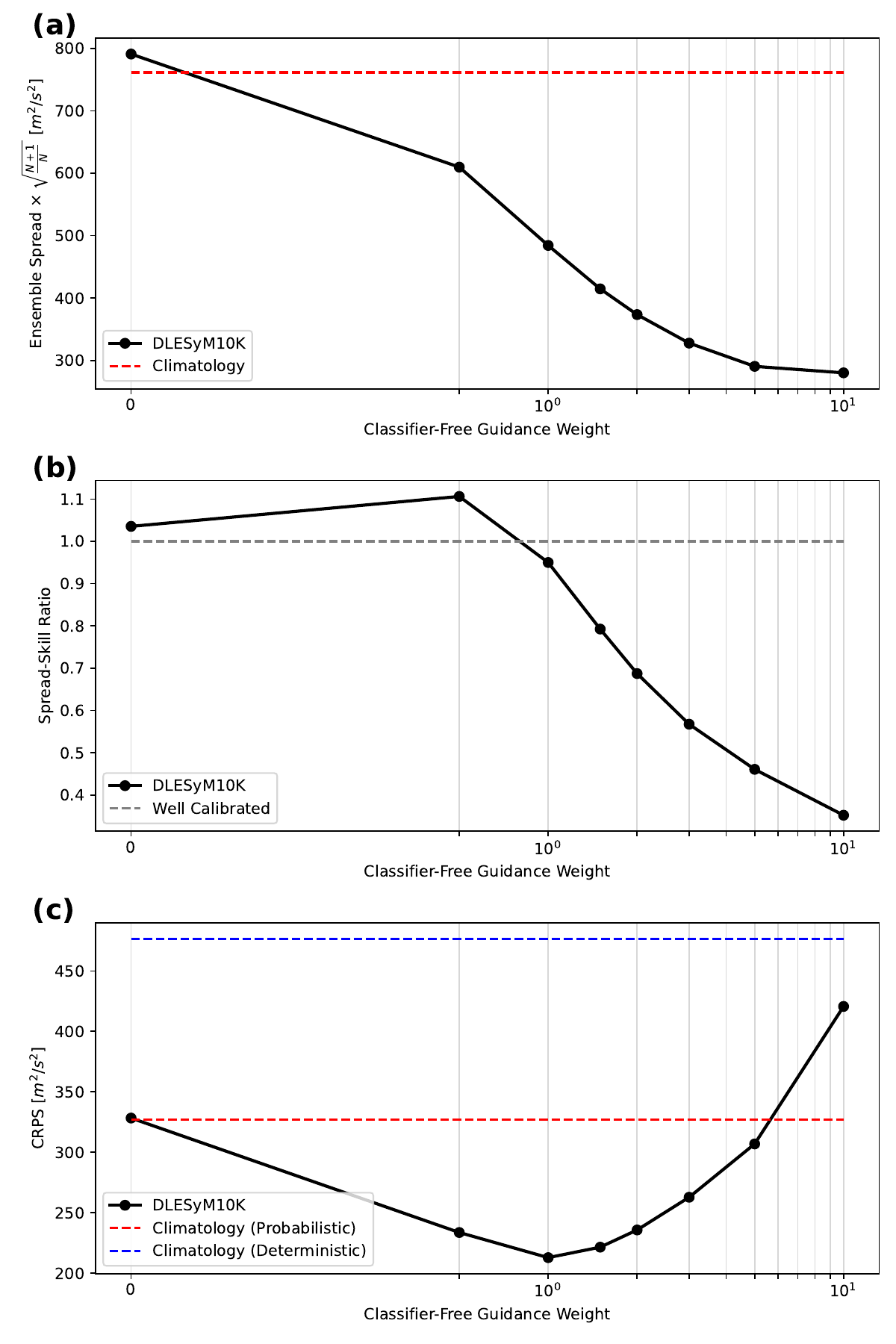}
    \caption{Calibration of medium-range distilled student model forecasts using classifier-free guidance. (a) Day 7 global $Z_{500}$ ensemble spread for varying classifier-free guidance strengths (black solid) with reference ensemble spread for a 20 year climatology (red dashed). (b) Distilled model ensemble spread-skill ratio for varying classifier-free guidance strengths (black solid). Ensemble spread-skill plot uses unbiased estimator for ensemble spread. (c) Day 7 global mean $Z_{500}$ CRPS for distilled student model with varying classifier-free guidance strengths (black solid), compared to a probabilistic (red dashed) and deterministic (blue dashed) 20 year climatology. Vertical grid lines align with integer guidance strengths, with an extra line at 0.5 between 0 and 1. Each point on the plot is averaged over 400 initialization dates (Section \ref{methods:model_exp}) with ensemble size of 32 (Table \ref{tab:sampler_settings}).}
    \label{fig:cfg_spread}
\end{figure}

The ensemble spread of our medium-range (7-day lead time) DLESyM10K forecasts varies monotonically with classifier-free guidance strength, enabling simple calibration (Figure \ref{fig:cfg_spread}). When the guidance strength is set to zero, the model samples unconditionally and thus its spread is close to that of climatology. In this limit the model is conditioned only on the day of year and simply generates samples from the learned climatology of its training data (panel a). As the guidance is increased, the ensemble spread drops as the model generation depends more strongly on the input conditioning. When the guidance strength is close to unity, the predictions achieve an optimal tradeoff between forecast spread and accuracy, with a spread-skill ratio of one and the minimum CRPS (panel b). In hindsight, further exploration of guidance strengths between 0.5 and 1.0 would have helped to find precisely the optimal guidance strength. As the guidance is increased beyond one, the ensemble spread continues to decrease but the ensemble RMSE increases, leading to an increasingly overconfident but inaccurate forecast and thus a low spread-skill ratio and high CRPS. Interestingly, here the optimal CRPS was achieved using classifier-free guidance strength of one (which is used throughout the rest of the manuscript), suggesting that the training objective of the conditional model could implicitly encourage a good spread-skill ratio even with no further tuning of classifier-free guidance strength. Nonetheless, in cases where simple conditional sampling (guidance strength of one) doesn't immediately yield a well calibrated forecast, we have shown here that adjusting the ensemble spread of the forecast can be achieved by simply changing the classifier-free guidance strength used at inference, providing a convenient method for calibrating ensemble forecasts post-training. 



\subsubsection{Distilled Models Are Skillful Across Lead Times}\label{results:model_exp:skill}



\begin{figure}[t]
    \centering
    \includegraphics[width = \columnwidth]{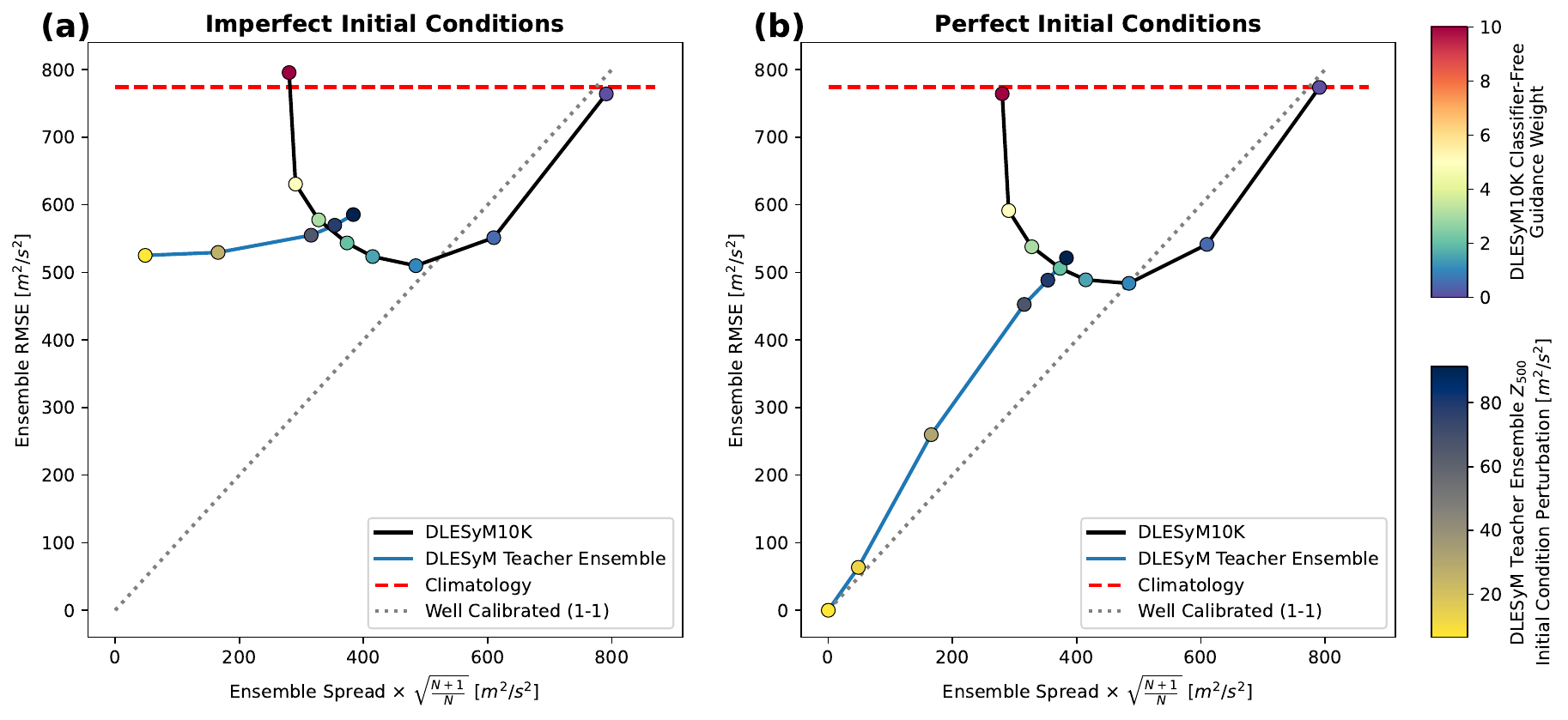}
    \caption{Spread-skill plot for day 7 $Z_{500}$ forecasts comparing distilled student models to initial condition ensembles from the autoregressive teacher model. (a) Spread-skill obtained given an imperfect initial condition (Section \ref{methods:model_exp}) using the distilled student model (solid black) and initial condition ensembles from the autoregressive teacher model (solid blue). The 1-1 calibration line (grey dashed) and climatology RMSE (red dashed) are shown for reference. For the distilled student model, the classifier-free guidance weight used is indicated by the color of the marker whereas for the autoregressive teacher it indicates initial condition perturbation strength used to generate the ensemble. (b) Same as (a) but for the idealized case of perfect initial conditions. Each point on the plot is averaged over 400 initialization dates (Section \ref{methods:model_exp}) with ensemble size of 32 (Table \ref{tab:sampler_settings}).}
    \label{fig:medium_spread_skill}.
\end{figure}

\textbf{Medium range.} We first assess medium-range (7-day) forecast skill under a realistic setting where both models are initialized from an imperfect estimate of the atmospheric state that contains non-negligible error (Figure \ref{fig:medium_spread_skill}a). In this setting, the autoregressive teacher model (DLESyM) exhibits substantial ensemble RMSE because initial condition errors grow during the autoregressive rollout (Figure \ref{fig:ic_error_growth}). Increasing ensemble spread by further perturbing around the already-imperfect initial condition does not meaningfully reduce forecast error, as the 7-day evolution remains tightly constrained by the biased starting point. In contrast, DLESyM10K shows much greater robustness to this initialization error: its day 7 $Z_{500}$ ensemble RMSE increases by only $25$~m$^2$/s$^2$ due to the addition of initial condition error (comparing panels a \& b), and its RMSE is lower than that of the autoregressive teacher for moderate classifier-free guidance strengths. This improved performance stems from the student model’s probabilistic formulation, which represents a distribution of plausible future states and thereby expresses aleatoric model uncertainty that can hedge over multiple potential trajectories consistent with an imperfect initial condition. Equivalent benefits for the teacher model would require explicitly injecting model uncertainty, for example through a checkpoint ensemble~\cite{weyn2021sub} or probabilistic model design~\cite{lang2024aifs,alet2025skillful,bonev2025fourcastnet}. By varying the classifier-free guidance strength, the distilled student model is able to explore spread-skill space. When the guidance strength is $\sim$1 the distilled student model achieves a well-calibrated forecast by intersecting the 1-1 line, whereas the autoregressive teacher model is under-dispersive across the full range of initial condition perturbation strengths explored here.

It is instructive to also examine forecast behavior in the idealized regime of perfect initial conditions (Figure \ref{fig:medium_spread_skill}b). When initialized exactly from the true atmospheric state and run deterministically, the autoregressive teacher produces a perfect forecast with zero RMSE since it is a perfect model of the DLESyM atmosphere. As perturbations are added around the true state, its ensemble RMSE increases smoothly with perturbation amplitude. DLESyM10K behaves differently in this idealized setting: even when initialized from the true state, it yields a non-zero ensemble RMSE due to its probabilistic nature. Increasing perturbation amplitude does not significantly change its RMSE (comparing panels a \& b), reflecting the fact that uncertainty in DLESyM10K arises predominantly from its learned distribution over future states rather than from sensitivity to initial conditions.

\begin{figure}[t]
    \centering
    \includegraphics[width = 0.7\columnwidth]{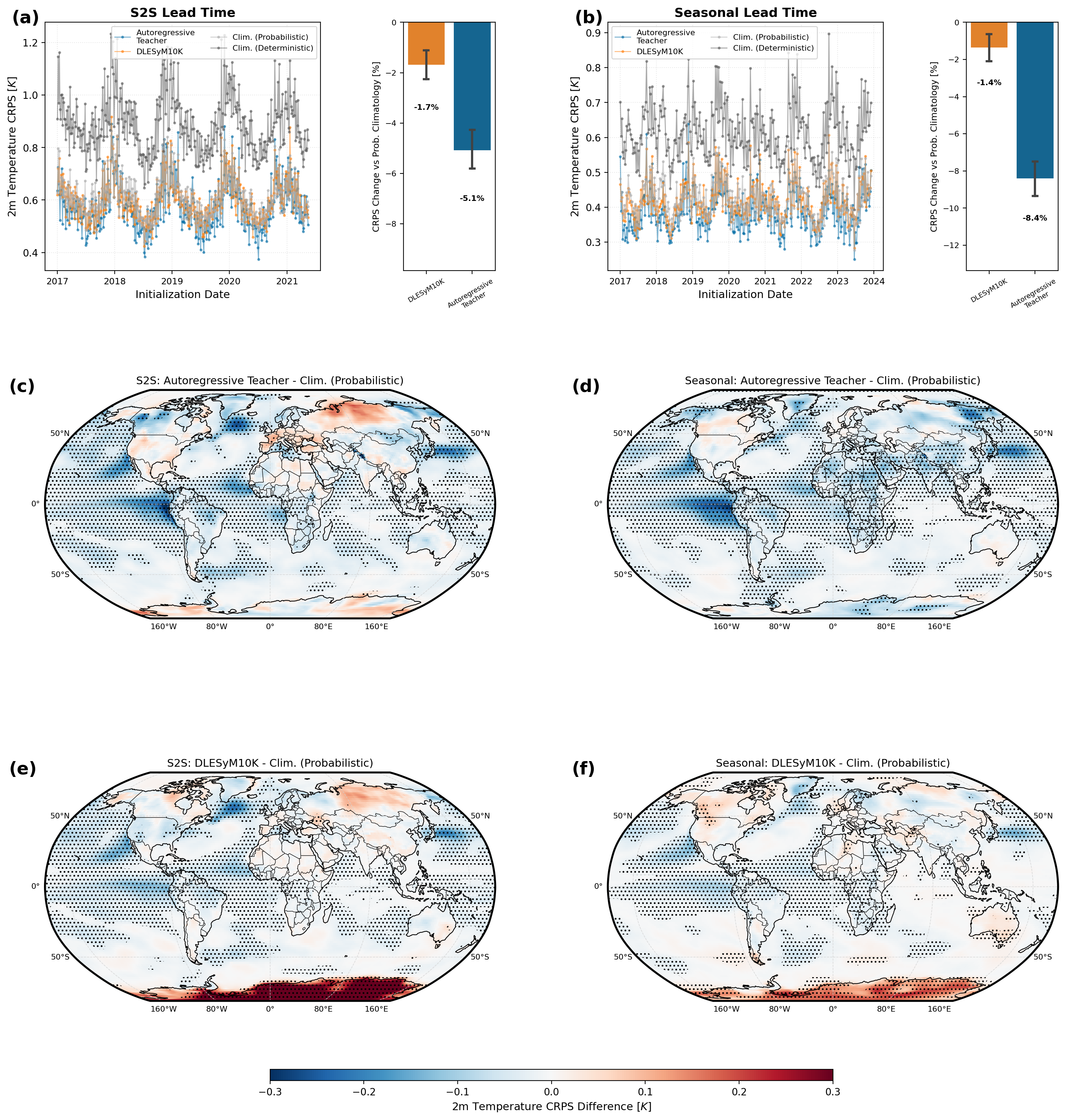}
    \caption{Distilled student model long-range forecast skill in perfect-model experiment. (a) Global mean 2-m air temperature CRPS at 4-week lead time for the autoregressive teacher (blue), distilled student (orange), deterministic climatology (light grey), and probabilistic climatology (dark grey) with the associated percentage change with respect to probabilistic climatology shown in the accompanying bars with error bars showing 95\% confidence interval by bootstrapping over evaluation dates. (b) Same as (a) but at 12-week lead time. (c) Global map of time mean difference between autoregressive teacher CRPS and that of probabilistic climatology at 4-week lead time. (d) Same as (c) but for 12-week lead time. (e) Same as (c) but for DLESyM10K and (f) same as (d) but for DLESyM10K. In (c-f) blue indicates an improvement with respect to climatology and stippling indicates a difference that is significantly different from zero at significance level 0.05 (\ref{appendix:significance_testing}).}
    \label{fig:long-range_model-world_skill}
\end{figure}

\textbf{S2S and seasonal.} Moving to long-range forecasting, we now evaluate the performance of DLESyM10K at the more challenging S2S and seasonal time-scales in the setting where we have an imperfect initial condition (Figure \ref{fig:long-range_model-world_skill}). At both S2S and seasonal lead times, DLESyM10K shows substantially more skill than a deterministic climatology (panels a \& b), achieving global mean skill within 3-7\% of an ensemble forecast using the autoregressive teacher model, constructed by further perturbing around the already imperfect initial conditions provided to DLESyM10K. Comparing both DLESyM10K and the autoregressive teacher to a probabilistic climatology highlights the significant challenge of long-range forecasting as they both only marginally improve upon probabilistic climatology in terms of global mean CRPS (bar plots in panels a \& b). Nonetheless, DLESyM10K shows statistically significant (\ref{appendix:significance_testing}) skill relative to climatology both in the global mean (panels a \& b) and in many regions (panels e \& f), with the model showing most skill in the tropics and over oceans where the relatively slow evolution of sea surface temperature increases the long-range predictability of surface air temperature, with similar spatial structure to that seen in the autoregressive teacher (panels c \& d). At both S2S and seasonal lead times DLESyM10K appear to struggle primarily over the continental Northern Hemisphere and over Antarctica where there appear to be significant errors that warrant closer scrutiny in future work. Even with these limitations, DLESyM10K exceeds climatology and closely matches the skill of its autoregressive teacher in a single model timestep, which corresponds to 126 and 392 teacher steps at S2S and seasonal timescales respectively.


\section{Real-World S2S Forecasting}

\subsection{Setup}\label{real_world_methods}

\subsubsection{Fine-Tuning Distilled Models on ERA5}\label{methods:era5_finetuning}

Ultimately, our goal is to apply DLESyM10K to real-world long-range forecasting taking ERA5 reanalysis as initial condition. While DLESyM was trained on ERA5, the model climate that emerges when it is run for long periods does not exactly match that of the real-world \cite{cresswell2025deep}, as is the case for any climate model. If we are to apply DLESyM10K, which was trained on DLESyM simulations, for real-world forecasting there is thus a domain shift to overcome. 

To address this domain shift, we fine-tune DLESyM10K on ERA5 data. During fine-tuning, we reduce the learning rate by a factor of ten compared to the original training and freeze all weights except for: the first convolutional layer in the UNet encoder, the final UNet block in the decoder, all UNet blocks with attention (which are near the middle of the UNet), and all GroupNorm weights and biases. During early testing, we found this strategy to achieve a good balance, providing enough capacity to adapt to ERA5 data while freezing enough of the pre-trained weights to avoid overfitting to the short ERA5 record.

\subsubsection{Bias Correcting Long-Range Forecasts Using Lead Time-Dependent Model Climatology}\label{methods:bias_correction}

For both physics-based and autoregressive weather models it is common practice to perform bias correction when forecasting at S2S and seasonal time-scales to correct for model drift caused by accumulating model errors \cite{vitart2004monthly,weigel2008probabilistic,vitart2017subseasonal,weyn2021sub,weyn2024ensemble}. In this study we use a lead time-dependent model climatology to bias correct all real-world long-range forecasts. In this approach, we take reforecasts for the same day of the year for the past 20 years and find the mean bias of the reforecasts compared to ERA5 ground truth. This mean bias is then subtracted from the forecast fields for the target year before scoring. Note while DLESyM10K does not in theory suffer an accumulation of errors over long autoregressive rollouts, we found in practice the same bias correction procedure to be helpful for our distilled models when applied to real-world data. This is likely a result of the DLESyM-ERA5 domain shift as bias correction was not found to be beneficial in our perfect-model experiment.

\subsubsection{Long-range distillation with a reduced order student model}

Predictability at S2S timescales is notoriously weak, and the slow, largely linear modes of variability that dominate at these timescales mean that reduced-order linear methods like linear inverse models (LIM) \cite{penland1995optimal,newman2003study,albers2021subseasonal} can remain competitive with far more expressive models. We therefore explore whether the benefits of our long-range distillation framework, pre-training on large synthetic data followed by real-world fine-tuning, extend to a LIM student model and benchmark these LIM models against real world S2S forecasting baselines and the diffusion models.

\textbf{Model.} We fit a discrete LIM for 2\,m temperature anomalies in a truncated EOF subspace. Let $x' \in \mathbb{R}^{d}$ denote the input anomaly, the 7-day mean ending at initialization, and $y' \in \mathbb{R}^{d}$ the target anomaly, the 7-day mean centred 28 days after initialization, both on the $d$-pixel HEALPix64 grid and formed relative to a daily climatology. 

With covariances $\mathbf{C}_{xx}=\mathbb{E}[x'x'^{\top}]$, $\mathbf{C}_{yx}=\mathbb{E}[y'x'^{\top}]$ and $\mathbf{C}_{yy}=\mathbb{E}[y'y'^{\top}]$, the LIM forecast is the best linear predictor of $y'$ given $x'$,
\begin{equation}
    \hat{y}' = \mathbf{G}x', \qquad \mathbf{G} = \mathbf{C}_{yx}\mathbf{C}_{xx}^{-1}.
    \label{eq:lim_G_pop}
\end{equation}
If $x'$ and $y'$ are jointly Gaussian, this is the conditional mean, and the forecast distribution is $y' \mid x' \sim \mathcal{N}(\mathbf{G}x', \boldsymbol{\Sigma})$ with
\begin{equation}
    \boldsymbol{\Sigma} = \mathbf{C}_{yy} - \mathbf{C}_{yx}\mathbf{C}_{xx}^{-1}\mathbf{C}_{yx}^{\top}.
    \label{eq:lim_Sigma_pop}
\end{equation}

\textbf{Estimation in a reduced-rank EOF basis.} In practice the covariances must be estimated from $n$ training pairs, e.g. $\hat{\mathbf{C}}_{xx} = \frac{1}{n}\sum_{i} x'_i x_i'^{\top}$ and $\hat{\mathbf{C}}_{yx} = \frac{1}{n}\sum_{i} y'_i x_i'^{\top}$. This is not tractable in data space on the full grid which has $d \approx 5\times10^{4}$ pixels and ERA5 provides only a few thousand samples. We therefore estimate the LIM in the subspace spanned by the leading $K$ EOFs. Let $\mathbf{E}\in\mathbb{R}^{d\times K}$ hold these as orthonormal columns and $z=\mathbf{E}^{\top}x'$ denote the EOF coefficients. The propagator is estimated from the $K\times K$ coefficient covariances and mapped back to physical space:
\begin{equation}
    \hat{\mathbf{G}}_{z} = \hat{\mathbf{C}}^{z}_{yx}\left(\hat{\mathbf{C}}^{z}_{xx}\right)^{-1},
    \qquad
    \hat{\mathbf{G}} = \mathbf{E}\,\hat{\mathbf{G}}_{z}\,\mathbf{E}^{\top}.
    \label{eq:lim_G}
\end{equation}
The truncation thus regularizes the covariance estimates, retaining the well-sampled, high-variance directions and discarding the rest. In practice, the EOF basis itself must also be estimated from finite training samples, $n$. To enable transfer between DLESyM and ERA5 data in a common coordinate system, we compute a single EOF basis from the ERA5 training data and use it as a common basis. Further details and justifications for this choice of EOF basis are provided in Appendix \ref{appendix:lim:basis}.


Given a fixed EOF basis, the fidelity of the LIM predictions is then governed by the accuracy with which we can estimate the sample covariances in EOF-space $\hat{\mathbf{C}}^{z}_{yx}$ and $\hat{\mathbf{C}}^{z}_{xx}$ from finite training data. 


\textbf{Forecast and probabilistic ensemble.} The LIM forecast in physical space is formed by 
\begin{align}
    \hat{y} &= \mathrm{clim}_{\mathrm{tgt}} + \hat{\mathbf{G}} x' \\  & = \mathrm{clim}_{\mathrm{tgt}} + \mathbf{E}\,\hat{\mathbf{G}}^z\,\mathbf{E}^\top x'.
\end{align}
To construct a probabilistic ensemble, we then apply climatological spread by sampling anomalies from the prior 20 years and adding these to $\hat{y}$. This approach matches that taken in our probabilistic climatology baseline, so improvements in CRPS for the LIM are driven solely by improvements in the ensemble mean. The rationale for this ensembling strategy rather than using the LIM's Gaussian posterior covariance is discussed in Appendix \ref{appendix:lim:ensemble}.

\begin{figure}[t]
    \centering
    \includegraphics[width = 0.8\columnwidth]{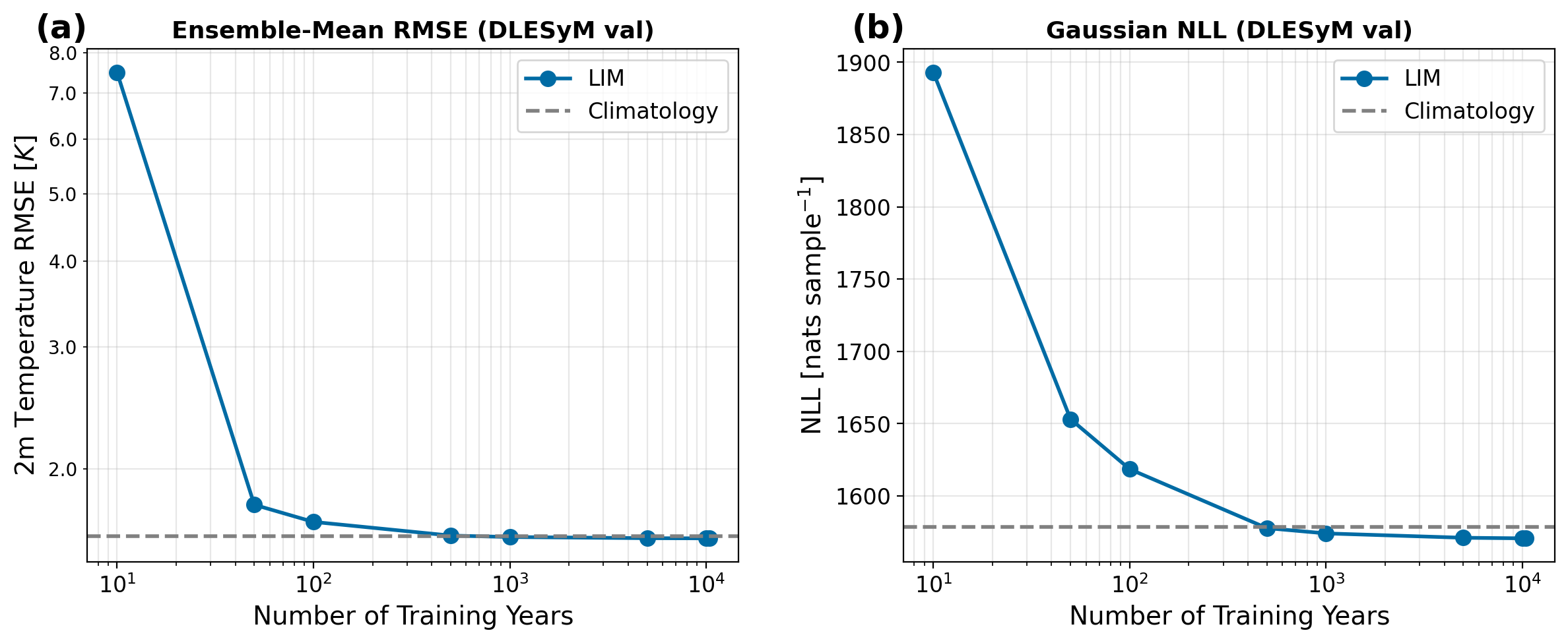}
    \caption{Quality of the LIM fit improves with increasing volume of synthetic data. (a) Validation MSE of 4-week 2\,m temperature forecasts on withheld DLESyM data, as a function of the number of training years; the dashed line shows climatology. (b) Same as (a) but for the predictive negative log-likelihood, $-\log p(z_\tau \mid z_0)$, which scores both the forecast mean and the forecast-error covariance (Appendix \ref{appendix:lim:nll}). Both panels show results for estimating $\hat{\mathbf{G}}_{z,\mathrm{DLESyM}}$ with a fixed EOF truncation of $K=500$.}
    \label{fig:lim_years_ablation}
\end{figure}

\textbf{Synthetic vs. real-world training.} Given a fixed EOF basis, the LIM propagator, $\hat{\mathbf{G}}_z$, is constructed from finite data through Equation \ref{eq:lim_G} and can be fit either on real-world ERA5 data, $\hat{\mathbf{G}}_{\mathrm{ERA5}}$, or from the much larger synthetic DLESyM simulated data, $\hat{\mathbf{G}}_{\mathrm{DLESyM}}$. The limited sample size of ERA5 means that in practice we must severely truncate $K$ for $\hat{\mathbf{G}}_{\mathrm{ERA5}}$ to prevent overfitting. In contrast, the large synthetic dataset offers a path to estimating the higher-mode covariances and enabling a higher-rank $\hat{\mathbf{G}}_{\mathrm{DLESyM}}$. As for the diffusion student model
(Section~\ref{results:model_exp:scaling}), the quality of the LIM fit improves
steadily with the volume of synthetic data up to at least $\sim$1000 years
(Figure~\ref{fig:lim_years_ablation}) when evaluated through RMSE or through the Gaussian posterior log-likelihood which captures the fidelity of the full posterior forecast distribution (Appendix \ref{appendix:lim:nll}).



\textbf{Mixed LIM.} Analogously to fine-tuning the DLESyM-trained diffusion model (Section~\ref{methods:era5_finetuning}), we define a mixed LIM that is fit to ERA5 but regularized toward the simulation-trained propagator. We solve
\begin{equation}
    \min_{\mathbf{G}_{z}}\;
    \frac{1}{n}\sum_{i=1}^{n}\left\Vert \mathbf{G}_{z}\mathbf{E}^{\top}x'_i
        - \mathbf{E}^{\top}y'_i\right\Vert^{2}
    + \lambda\left\Vert \mathbf{G}_{z} - \hat{\mathbf{G}}_{z,\mathrm{DLESyM}}\right\Vert_{F}^{2},
    \label{eq:lim_mixed_objective}
\end{equation}
where the sum runs over the ERA5 training pairs. The minimizer is
\begin{equation}
    \hat{\mathbf{G}}_{z,\mathrm{mixed}}
    = \left(\hat{\mathbf{C}}^{z}_{yx} + \lambda\,\hat{\mathbf{G}}_{z,\mathrm{DLESyM}}\right)
      \left(\hat{\mathbf{C}}^{z}_{xx} + \lambda\mathbf{I}\right)^{-1},
    \label{eq:lim_mixed}
\end{equation}
with $\hat{\mathbf{C}}^{z}$ the ERA5 coefficient covariances. We parameterize the scalar regularization strength as
\begin{equation}
    \lambda = \bar{c}\,\frac{1-\alpha}{\alpha},
    \qquad
    \bar{c} = \frac{1}{K}\operatorname{tr}\hat{\mathbf{C}}^{z}_{xx},
    \label{eq:lim_lambda}
\end{equation}
with $\alpha\in(0,1]$, so that $\alpha\to0$ recovers
$\hat{\mathbf{G}}_{\mathrm{DLESyM}}$ and $\alpha=1$ recovers
$\hat{\mathbf{G}}_{\mathrm{ERA5}}$. 

\textbf{Model selection and metrics.} We select hyperparameters on a withheld cross-validation period (2016--2018) that does not overlap with the final evaluation period. For $\hat{\mathbf{G}}_{\mathrm{ERA5}}$ and $\hat{\mathbf{G}}_{\mathrm{DLESyM}}$ we sweep over $K$, and for $\hat{\mathbf{G}}_{\mathrm{mixed}}$ we sweep jointly over $K$ and $\alpha$, selecting the configuration that minimizes the global mean MSE of the 2\,m temperature forecast in physical space. As a secondary diagnostic we report the posterior negative log-likelihood (NLL) of the Gaussian forecast distribution (Eq.~\ref{eq:lim_Sigma_pop}), relative to climatology (Appendix \ref{appendix:lim:nll}). Whereas the MSE is dominated by the high-variance leading modes, the NLL weights all modes equally in logarithmic terms, and so mainly reflects how well the forecast-error covariance is estimated in the many low-variance modes.

\subsubsection{Benchmarking Distilled Models Against Operational S2S Forecasts}\label{methods:real_world_s2s_exp}

We evaluate the ability of the distilled models (both the diffusion model, DLESyM10K, and the LIM, $\hat{\mathbf{G}}_{mixed}$) to do S2S forecasting in the real world. As in Section \ref{methods:model_exp}, we seek to predict a weekly average at 4-week lead time. To benchmark our approach against existing operational forecasts, we use the ECMWF S2S ensemble forecasts which contain 50 ensemble members \cite{vitart2017subseasonal}. Specifically, we evaluate the ECMWF real-time forecasts for the period 2018-2022 taking ERA5 to be the target. We bias correct (Section \ref{methods:bias_correction}) all ECMWF forecasts using the reforecasts provided for the preceding twenty years using the same model version as the real-time forecasts and we bias correct DLESyM10K using the same procedure. DLESyM10K is fine-tuned on ERA5 data from 1980-2016. To test the value of distilling the large DLESyM simulation, we also evaluate a long timestep model with the same architecture trained from scratch on ERA5. Analogously, we assess the value of distillation in the LIM student models by comparing $\hat{\mathbf{G}}_{ERA5}$, $\hat{\mathbf{G}}_{DLESyM}$, and $\hat{\mathbf{G}}_{mixed}$. All forecasts are evaluated using CRPS focusing on 2 m air temperature. As in Section \ref{methods:model_exp}, all forecasts are benchmarked against a deterministic and probabilistic climatological forecast taken here from the 20 preceding years of ERA5.



\subsection{Results: Synthetic Pre-Training Makes Distilled Models Competitive for Real-World S2S Forecasting}\label{results:real_exp}

\begin{figure}[t]
    \centering
    \includegraphics[width = 0.8\columnwidth]{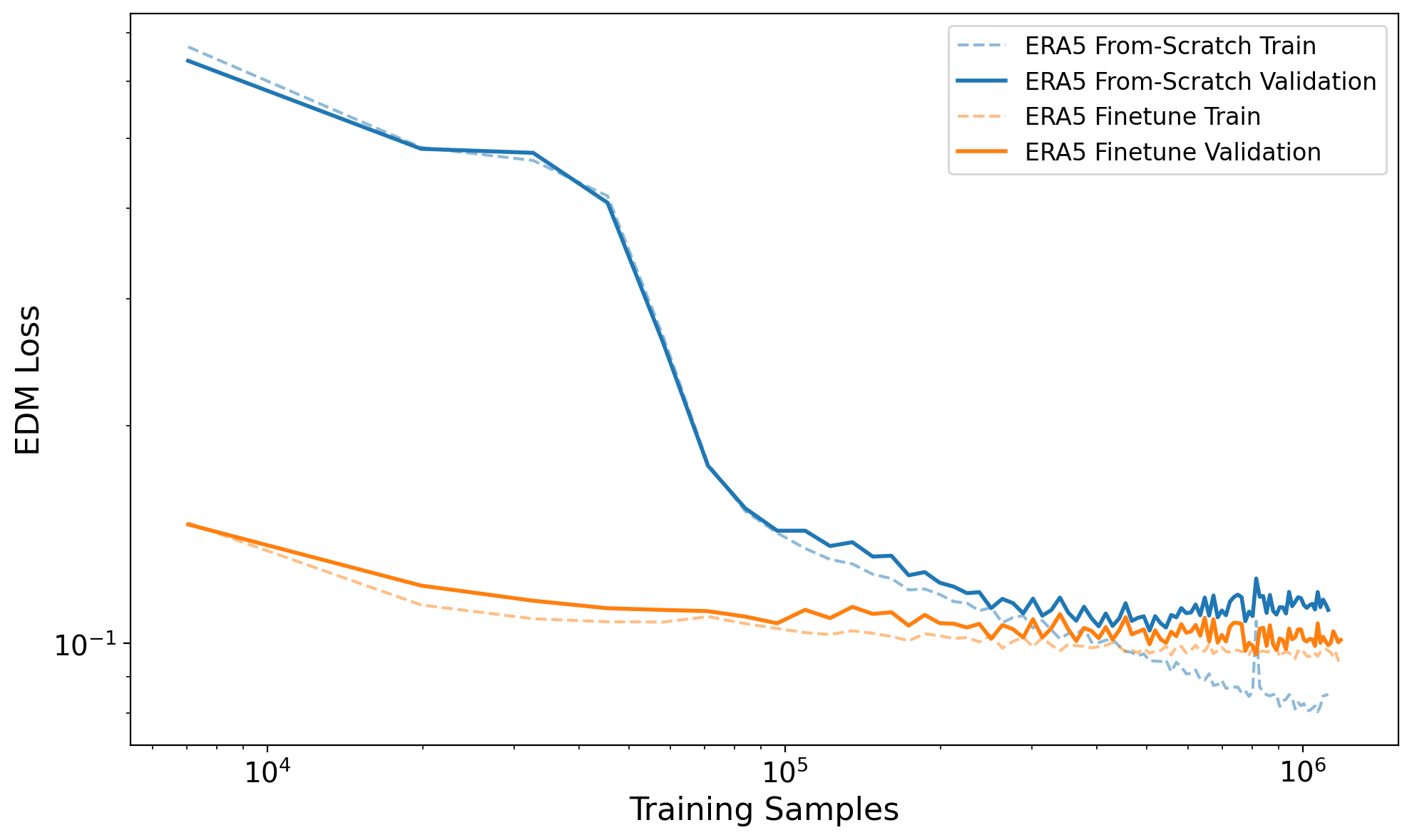}
    \caption{Fine-tuning distilled student models on real-world ERA5 reanalysis data for S2S forecasting. Orange curves: training (dashed) and validation (solid) loss for the DLESyM pre-trained model during fine-tuning on ERA5. Blue curves: same as orange but for model trained from scratch on ERA5. Note, the learning curves were smoothed for plotting purposes.}
    \label{fig:era5_scratch_v_finetune}
\end{figure}

The benefit of pre-training on synthetic data is reflected in the real world performance of both the diffusion and LIM student models. As outlined in Section \ref{methods:era5_finetuning}, we fine-tune the diffusion model DLESyM10K on real-world data from ERA5 to allow it to adapt to distribution shifts between the DLESyM simulated climate and the real world. Comparing the learning curves of our fine-tuned model to one with identical architecture but trained from scratch on ERA5 data (Figure \ref{fig:era5_scratch_v_finetune}) reveals the benefit of pre-training on DLESyM simulations. Training from scratch on ERA5 leads to overfitting, whereas fine-tuning only a subset of pre-trained parameters on ERA5 acts as a strong regularization, preventing overfitting. Importantly, the fine-tuned model is able to achieve a lower validation loss than training from scratch on ERA5, suggesting transferability of the skill learned from the DLESyM simulations to real-world data. 

\begin{figure}[t]
    \centering
    \includegraphics[width = 0.8\columnwidth]{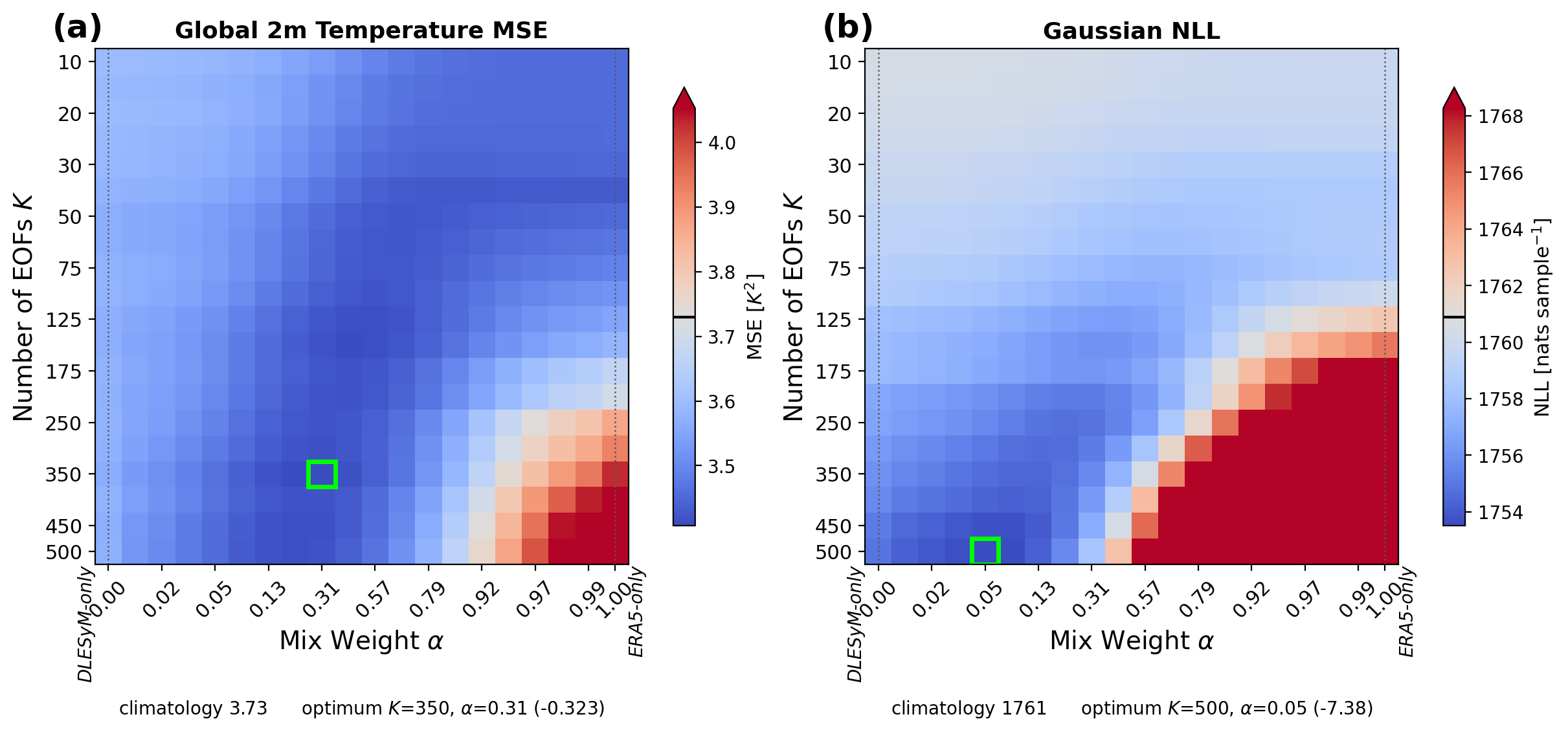}
    \caption{Mixed DLESyM-ERA5 LIM parameter sweep. (a) Global mean week-4 T2M MSE on 2016-2018 cross-validation years for varying EOF truncations, $K$, and dataset mixing weights, $\alpha$. (b) Same as (a) but for the Gaussian posterior negative log likelihood (defined in Appendix \ref{appendix:lim:nll}). In both panels, blue indicates improvement with respect to climatology and the green square highlights the optimal parameter combination.}
    \label{fig:lim_k-alpha_heatmaps}
\end{figure}

For the LIM student model, the benefit of synthetic pre-training is evident in the fact that the ($K$, $\alpha$) combination for $\hat{\mathbf{G}}_{mixed}$ with the minimum MSE lies at an intermediary $\alpha=0.31$ with a relatively high EOF dimensionality, $K=350$ (Figure \ref{fig:lim_k-alpha_heatmaps}a). The fidelity of the higher-order EOF modes in the Gaussian posterior forecast distribution, beyond just the MSE of the mean, is apparent in the fact that the Gaussian posterior NLL (Appendix \ref{appendix:lim:nll}) also decreases to a minimum at intermediary $\alpha$ and high $K$ (panel b). In contrast, the optimal ERA5-only model ($\alpha=1$) has much smaller dimensionality ($K=50$), with overfitting preventing accurate estimation of the higher mode covariances from the limited samples in ERA5. Distilling the large synthetic dataset thus allows estimating higher-rank covariances than from reanalysis alone, enabling more skillful LIM models than could be obtained from ERA5 alone.

\begin{figure}[t]
    \centering
    \includegraphics[width = \columnwidth]{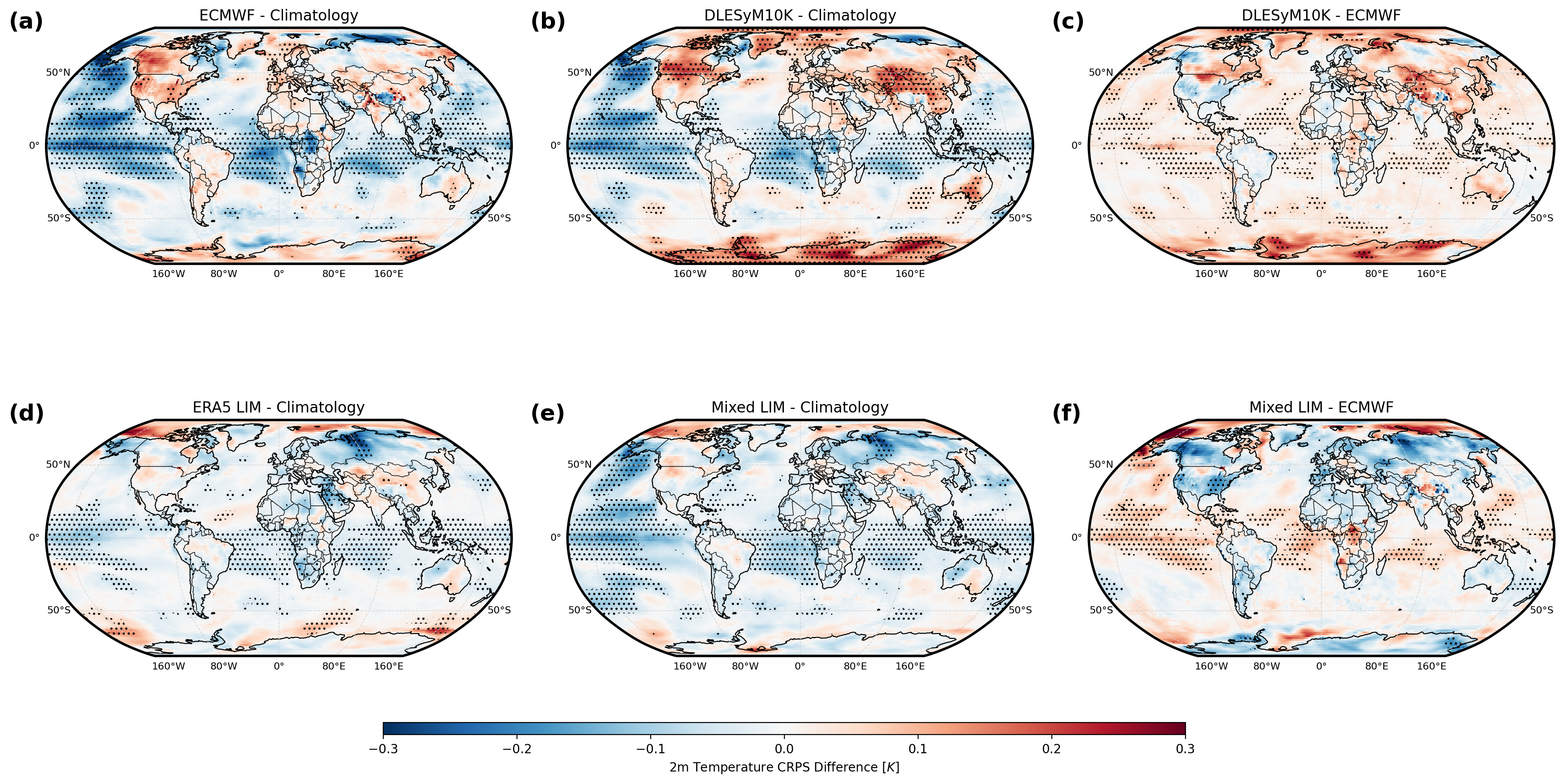}
    \caption{Comparison of the spatial patterns of real-world 4-week forecast skill between ECMWF and distilled student models for 2 m air temperature. (a) Difference between ECMWF and 20-year probabilistic climatology CRPS. (b) Same as (a) but for distilled student model. (c) Difference between distilled student model and ECMWF CRPS. (d-e) Same as (a) but for the linear inverse models (d) $\hat{\mathbf{G}}_{\mathrm{ERA5}}$ and (e) $\hat{\mathbf{G}}_{\mathrm{mixed}}$. (f) Difference between $\hat{\mathbf{G}}_{\mathrm{mixed}}$ and ECMWF CRPS. In all panels blue indicates an improvement in forecast skill and stippling indicates a statistically significant difference between the skill of the two methods under comparison (\ref{appendix:significance_testing}).}
    \label{fig:realworld_score_maps}
\end{figure}

To benchmark the distilled models against an operational baseline, we compare the week-4 2 m temperature forecast skill to that of the ECMWF IFS ensemble forecast system from 2018-2022 (Table \ref{tab:real_world_regional_crps} \& Figure \ref{fig:realworld_score_maps}). First, the challenge of forecasting at S2S time-scales is highlighted by the fact that all methods offer at best modest gains compared to a probabilistic climatology, taking the preceding 20 years as ensemble members, with skill mostly confined to the Tropics and over oceans. Nevertheless, the bias-corrected ECMWF forecasts, the ERA5-only LIM, and the mixed-training LIM all show statistically significant improvements over climatology (Table \ref{tab:real_world_regional_crps}) over land when averaged in the tropical band (30S-30N), with the mixed LIM showing the largest improvement with respect to climatology. For both the LIM and the diffusion model, the model that was pre-trained on DLESyM outperforms its counterpart trained only on ERA5, demonstrating the value of pre-training on a large synthetic training dataset for direct long-lead forecasting. The largest improvements with respect to climatology are found over oceans, where all methods outperform climatology, with predictability likely attributable to the relatively slow evolution of sea surface temperature, though all methods offer skill beyond a damped persistence baseline (Table \ref{tab:real_world_regional_crps}, \ref{appendix:baselines}). The fact that the mixed LIM outperforms the diffusion student models highlights that for the S2S prediction problem, EOF-regression training objectives may be better suited than full-state vector diffusion. Comparing our top-performing distilled model, the mixed LIM, to the operational state-of-the-art baseline, ECMWF IFS, the distilled model exhibits modest but statistically significant improvements in CRPS compared to ECMWF over N America and Eurasia, while under-performing ECMWF in Central Africa and over the tropical oceans (Figure \ref{fig:realworld_score_maps}f).

\begin{table}[t]
\centering
\small
\caption{Regional-mean 2\,m temperature skill, $S$, as percent improvement in fair CRPS over probabilistic climatology: $S = 100 \times \overline{(\mathrm{CRPS}_{\mathrm{clim}} - \mathrm{CRPS}_{\mathrm{fcst}})}\,/\,\overline{\mathrm{CRPS}_{\mathrm{clim}}}$, with overbars denoting means over initialization dates. Positive values beat climatology. Tropics are $|\mathrm{lat}| \le 30^\circ$ and extratropics $|\mathrm{lat}| > 30^\circ$. For bias-corrected methods, values in parentheses are without bias correction. An asterisk marks entries whose mean CRPS difference from climatology has a 95\% confidence interval that excludes zero, using a paired $t$-test with lag-1 effective sample size $N_{\mathrm{eff}}=N(1-r_1)/(1+r_1)$ on the time series of differences. For bias-corrected methods, the unbiased improvement is shown with the biased improvement in parentheses.}
\label{tab:real_world_regional_crps}
\begin{tabular}{lccc}
\toprule
\multicolumn{4}{l}{\textit{Land}} \\
Method & Global & Tropical & Extratropical \\
\cmidrule(lr){1-4}
ECMWF & -0.7 (-19.3$^{*}$) & 2.5$^{*}$ (-46.4$^{*}$) & -1.6$^{*}$ (-10.8$^{*}$) \\
Long Timestep ERA5 & -6.8$^{*}$ (-37.8$^{*}$) & 0.7 (-60.0$^{*}$) & -9.2$^{*}$ (-30.9$^{*}$) \\
DLESyM10K & -2.9 (-11.1$^{*}$) & 1.2 (-4.9$^{*}$) & -4.2 (-13.0$^{*}$) \\
Damped Persistence & -9.5$^{*}$ & -10.2$^{*}$ & -9.3$^{*}$ \\
DLESyM LIM & -0.2 & 0.1 & -0.3 \\
ERA5 LIM & 2.2 & 4.4$^{*}$ & 1.6 \\
Mixed LIM & 2.5 & 4.5$^{*}$ & 1.9 \\
\midrule
\multicolumn{4}{l}{\textit{Ocean}} \\
Method & Global & Tropical & Extratropical \\
\cmidrule(lr){1-4}
ECMWF & 10.9$^{*}$ (7.1$^{*}$) & 24.0$^{*}$ (16.4$^{*}$) & 4.6$^{*}$ (2.6$^{*}$) \\
Long Timestep ERA5 & 2.8$^{*}$ (-122.4$^{*}$) & 17.2$^{*}$ (-233.0$^{*}$) & -4.2$^{*}$ (-69.1$^{*}$) \\
DLESyM10K & 6.1$^{*}$ (-1.0) & 17.8$^{*}$ (10.9$^{*}$) & 0.5 (-6.6$^{*}$) \\
Damped Persistence & -4.3$^{*}$ & 2.1 & -7.3$^{*}$ \\
DLESyM LIM & 0.4 & 2.6 & -0.7 \\
ERA5 LIM & 2.8$^{*}$ & 8.2$^{*}$ & 0.1 \\
Mixed LIM & 7.4$^{*}$ & 15.3$^{*}$ & 3.6$^{*}$ \\
\bottomrule
\end{tabular}
\end{table}


\section{Conclusions}\label{conclusion}


Achieving skillful long-range forecasts, from S2S to seasonal time-scales, remains one of the most challenging and socioeconomically consequential goals in AI-driven weather prediction. It remains to be seen whether the current paradigm of training autoregressive models on reanalysis data can achieve breakthrough skill in forecasting the slower modes of climate variability that drive S2S and seasonal predictability. 


In this study, we proposed a novel approach to long-range weather modeling. Rather than relying on ensemble autoregressive rollouts, we repurpose a reanalysis-trained autoregressive model as a generator of large volumes of synthetic data to train a separate distilled model that predicts long-range forecasts in a single model timestep. This design leverages the ability of autoregressive models to produce realistic weather trajectories while mitigating their limitations for ensemble forecasting at extended lead times. To our knowledge, this is the first work to train an AI weather model on synthetic data generated by another AI model, although prior studies have trained on physics-based simulations \cite{rasp2021data}. Our results show that replacing long autoregressive rollouts with a single extended timestep and optimizing for a probabilistic long-range objective yields well-calibrated forecasts across multiple time scales. We further demonstrate that classifier-free guidance, commonly used in image generation, offers a simple recipe for controlling ensemble spread in conditional diffusion models. Training with such long timesteps (equivalent to up to 392 autoregressive steps) without overfitting required a massive synthetic dataset, and we observed that S2S forecast skill scaled with the number of synthetic years, even when the dataset far exceeded the size of ERA5. Fine-tuning on ERA5 reanalysis allowed our distilled models to achieve real-world forecast skill comparable to the operational ECMWF S2S ensemble forecast system, albeit with both systems showing limited skill beyond climatology over land. Finally, we demonstrated that the benefit of distilling synthetic data extends beyond AI models to classical data-driven approaches like linear inverse models, with the large synthetic training dataset enabling accurate covariance estimation for significantly higher-rank modes of variance than is achievable from the short reanalysis record alone, leading to enhanced forecast skill.


Although our findings highlight the potential of distillation techniques for improving long-range weather prediction, several limitations remain that warrant attention in future research. Firstly, while it is impressive that our model achieves competitive S2S forecast skill compared to ECMWF, our diffusion model still slightly trailed ECMWF in terms of global mean CRPS, while our distilled LIM model had only limited regions of statistically signifcant skill gain over ECMWF IFS. Long-range distillation thus does not yet represent a breakthrough in real-world S2S forecast skill. Note that S2S time-scales are notoriously challenging for both physics-based and data-driven weather models, and the current state of the art for AI S2S weather models is parity with, or at best marginal improvements upon, the ECMWF IFS ensemble forecasts \cite{chen2024machine,weyn2024ensemble,lang2024aifs} and our results underline that traditional data-driven approaches like LIMs remain competitive for S2S forecasting. We hope that future refinements to the long-range distillation approach outlined here could eventually yield more substantial improvements in S2S forecasting. 

The primary opportunity we see for improving the performance of long-range distilled models is by leveraging the rapid recent advances in coupled autoregressive Earth System Models. One benefit of our distillation approach, is that it can be readily applied to any autoregressive model, allowing us to bootstrap future refinements to the autoregressive teacher model into improved distilled models. While the version of DLESyM used in this study generates realistic interrannual and intraseasonal variability \cite{cresswell2025deep}, it simulates only 8 atmospheric variables and only a single ocean variable. It is thus likely that the synthetic climate our distilled models were trained on here does not fully capture important modes of S2S and seasonal variability in the real world such as the Madden-Julian Oscillation and ENSO, indeed the ENSO amplitude was observed to be weak compared to observations \cite{cresswell2025deep}. As increasingly sophisticated autoregressive Earth System Models emerge, we anticipate that applying our framework to models such as SamudrACE \cite{duncan2025samudrace,watt2025ace2,dheeshjith2025samudra}, NeuralGCM \cite{kochkov2024neural}, and forthcoming enhancements to DLESyM could further improve the forecast skill of long-range distilled models. By proposing training a separate model for long-range forecasting, we highlight that autoregressive climate emulators can be useful as generators of huge synthetic training datasets for long-range forecasting even if they are not themselves directly amenable to ensemble forecasting from reanalysis. Finally, the ERA5 fine-tuning method for distilled models presented here provides a crucial tool for ensuring the resulting forecast models are well calibrated against real-world observations while retaining the benefits of pre-training on multi-centennial simulations.

This study provides the first demonstration that forecast performance can improve with increasing volumes of AI-generated synthetic training data (Figure \ref{fig:training_scaling_w_years}) and that these improvements transfer to downstream real-world forecasts after ERA5 fine-tuning (Figure \ref{fig:realworld_score_maps}). This suggests a new scaling axis for AI weather models, which have to date only been scaled in terms of model capacity, whereas results from large language modeling indicate that model capacity and training dataset size should be scaled in tandem \cite{kaplan2020scaling}. Here, S2S forecast skill scaled with diminishing returns as dataset size increased, rather than following a power law, but dataset size was varied at fixed model capacity. Future work could explore whether jointly scaling model capacity and dataset size, as is done for large language models, yields stronger gains, potentially benefiting from transformer-based architectures that scale more effectively with large datasets than the convolutional UNet backbone used here \cite{kaplan2020scaling}. While this study focused on long-range forecasting, training on huge synthetic datasets could also improve short- and medium-range forecasts, either by leveraging existing large ensemble simulations \cite{mahesh2025henspart1,mahesh2025henspart2} or by exploiting the low inference cost of AI models to generate training data from any teacher model of interest. Finally, this study showed that large simulations from autoregressive emulators can benefit statistical methods beyond deep learning that are limited by the length of the observational record; here, they allowed the LIM to estimate lagged covariances across many more EOF modes than reanalysis alone supports. This suggests inferencing pre-trained AI emulators can serve domain scientists as inexpensive generators of large synthetic datasets for scientific and statistical analysis, without requiring them to train their own bespoke machine learning models.

By an initial exploration of distilled AI models trained on large volumes of synthetic data from autoregressive AI weather models, we hope to expand the design space of AI weather modeling and inspire further exploration of synthetic training data approaches. Departing from the autoregressive modeling framework opens up a broader range of climate informatics and scientific exploration applications beyond forecasting. \citeA{brenowitz2025climate} showed that training conditional generative models on ERA5 and ICON simulations outside of an autoregressive framework enabled a diverse range of use cases including downscaling, channel in-filling, and steerable sampling. While in this study, our distilled models were tasked with long-range forecasting, one could envisage training similar conditional models on huge autoregressive AI simulations for other tasks such as exploring the global drivers of local weather extremes or for conditioning sample generation on large-scale climate indices or other quantities of interest. 

\appendix
\section{Difference in Scores Significance Testing}\label{appendix:significance_testing}

Throughout this manuscript where we plot maps of differences in scores between one forecasting model and another, we indicate statistical significance using a paired test on the score differences, in a similar spirit to
\citeA{mardani2025residual}. The purpose of this test is to assess whether observed differences between the time-mean scores of two models are statistically significant given the large day-to-day variability of the atmosphere, and hence of the scores.

At each grid point we form the time series of paired score differences $d_t = s^{(A)}_t - s^{(B)}_t$ between models A and B. Because successive initialization dates are temporally correlated, treating the nominal sample size $N$ as independent would overstate significance. We therefore estimate an effective sample size from the lag-1 autocorrelation $r_1$ of $\{d_t\}$ (i.e.\ $\mathrm{Corr}(d_t,d_{t+1})$),
\begin{equation}
N_{\mathrm{eff}} = N\,\frac{1-r_1}{1+r_1},
\end{equation}
clipped to the interval $[2,N]$ for numerical stability, and perform a two-tailed $t$-test of whether the mean difference is zero using standard error $\mathrm{std}(d)/\sqrt{N_{\mathrm{eff}}}$ and $N_{\mathrm{eff}}-1$ degrees of freedom. Grid points with two-tailed $p$-values below $0.05$ are then retained after applying the Benjamini--Hochberg procedure to control the false discovery rate across grid points.

In Figure~\ref{fig:long-range_model-world_skill} we present bar plots of percentage improvement in global-mean scores with respect to probabilistic climatology,
with error bars. These error bars are obtained from the time series of percentage improvement (computed separately at each initialization), using a bootstrap estimate of the 95\% confidence interval over the time samples.

\section{Hyperparameters}\label{sec:hyperparams}

Details regarding the DLESyM ensemble simulation are provided in Table \ref{tab:dlesym_simulation_settings}. The training configurations for the distilled student model are in Table \ref{tab:hyperparams}. The diffusion sampler settings used for inference of the distilled student models are in Table \ref{tab:sampler_settings}.

\begin{table}
\caption{Details about the huge DLESyM ensemble simulation used to generate training data.}\label{tab:dlesym_simulation_settings}
\centering
\begin{tabular}{ll}
\hline
 Parameter  & Value  \\
\hline
  Model timestep & 6 hours \\
  Output frequency saved & Daily averages \\
  Data volume & 7 TB \\
  Number of stable ensemble members  & 162 \\
  Initialization date range  & 2008-01-01 to 2016-12-31 \\
  Simulation duration  & 90 years \\
  Spin-up time excluded  & First 2 years \\
  Training-Validation Split & 75\% for training, 25\% for validation \\
  Hardware run on  & 96 H100 GPUs \\
  Inference time  & 4 hours \\
  Simulation throughput  & 1,100 SYPD per H100 \\
\hline
\multicolumn{2}{l}{}
\end{tabular}
\end{table}

\begin{table}
\caption{Training configurations for distilled student models.}\label{tab:hyperparams}
\centering
\begin{tabular}{llll}
\hline
 Parameter  & DLESyM Training & ERA5 Fine-Tuning & ERA5 From Scratch  \\
\hline
  Distilled student architecture  & Song UNet HPX64 & Song UNet HPX64 & Song UNet HPX64   \\
  Number of parameters  & 149M & 149M & 149M   \\
  Batch size & 64 & 64 & 64   \\
  Learning rate & $10^{-4}$ & $10^{-5}$ & $10^{-4}$   \\
  Condition dropout fraction & 0.1 & 0.1 & 0.1   \\
  Optimizer & Adam & Adam & Adam   \\
  Number of training steps & 156,000 & 18,750 & 18,750 \\
  Training Time on 16 H100 GPUs & 20 hours & 2.4 hours & 2.4 hours \\
\hline
\multicolumn{4}{l}{}
\end{tabular}
\end{table}

\begin{table}
\caption{Diffusion sampler settings used for all distilled student forecasts presented in this study except where stated otherwise.}\label{tab:sampler_settings}
\centering
\begin{tabular}{l c}
\hline
 Parameter  & Value  \\
\hline
  $\sigma_{min}$  & 0.002 \\
  $\sigma_{max}$  & 200 \\
  Training $\sigma$ distribution  & Log uniform \\
  Number of sampler steps  & 18 \\
  $S_{churn}$ & 0 \\
  Sampler scheme & EDM ODE \\
  Classifier-free guidance strength & 1.0 \\
  Ensemble members & 32 \\
\hline
\multicolumn{2}{l}{}
\end{tabular}
\end{table}

\section{Supplementary Forecast Skill Evaluations}

Regional breakdown of 2\,m temperature CRPS scores in the model world experiment are provided in Table \ref{tab:model_world_regional_crps}

\begin{table}[t]
\centering
\caption{Regional mean 2\,m temperature CRPS [$K$] in the model world experiment. Scores are averaged over initialization dates, and using area-weighted averaging in space. Tropics are defined here as $|\mathrm{latitude}| \le 30^\circ$.}
\label{tab:model_world_regional_crps}
\begin{tabular}{lcccc}
\toprule
Region & Autoregressive Teacher & DLESyM10K & Clim.\ (Probabilistic) & Clim.\ (Deterministic) \\
\midrule
\multicolumn{5}{l}{\textit{S2S lead time}} \\
Global land only & 1.15 & 1.18 & 1.16 & 1.70 \\
Global ocean only & 0.31 & 0.32 & 0.34 & 0.50 \\
Tropical ocean & 0.18 & 0.19 & 0.22 & 0.32 \\
Tropical land & 0.60 & 0.62 & 0.63 & 0.91 \\
Extratropical ocean & 0.45 & 0.46 & 0.48 & 0.70 \\
Extratropical land & 1.58 & 1.63 & 1.58 & 2.31 \\
\midrule
\multicolumn{5}{l}{\textit{Seasonal lead time}} \\
Global land only & 0.73 & 0.78 & 0.77 & 1.11 \\
Global ocean only & 0.22 & 0.25 & 0.26 & 0.37 \\
Tropical ocean & 0.14 & 0.17 & 0.18 & 0.26 \\
Tropical land & 0.38 & 0.42 & 0.43 & 0.62 \\
Extratropical ocean & 0.31 & 0.33 & 0.34 & 0.50 \\
Extratropical land & 1.00 & 1.05 & 1.03 & 1.50 \\
\bottomrule
\end{tabular}
\end{table}

\section{Supplementary Analysis for "4.1.3: Long-range distillation with a reduced order student model"}\label{appendix:lim}

\subsection{Estimation of EOF basis}\label{appendix:lim:basis}

The EOF basis is computed by exact singular value decomposition of the ERA5 training anomalies. Because the HEALPix grid is equal-area, no area weighting is required. We retain $K_{\max}=500$ modes, which capture 96\% of the ERA5 training variance, and use this same basis for all LIMs. We estimate the basis from ERA5 rather than DLESyM because ERA5 is the target domain for real-world forecasting, and its EOFs are the patterns that explain the most variance there.

\begin{figure}[t]
    \centering
    \includegraphics[width=\columnwidth]{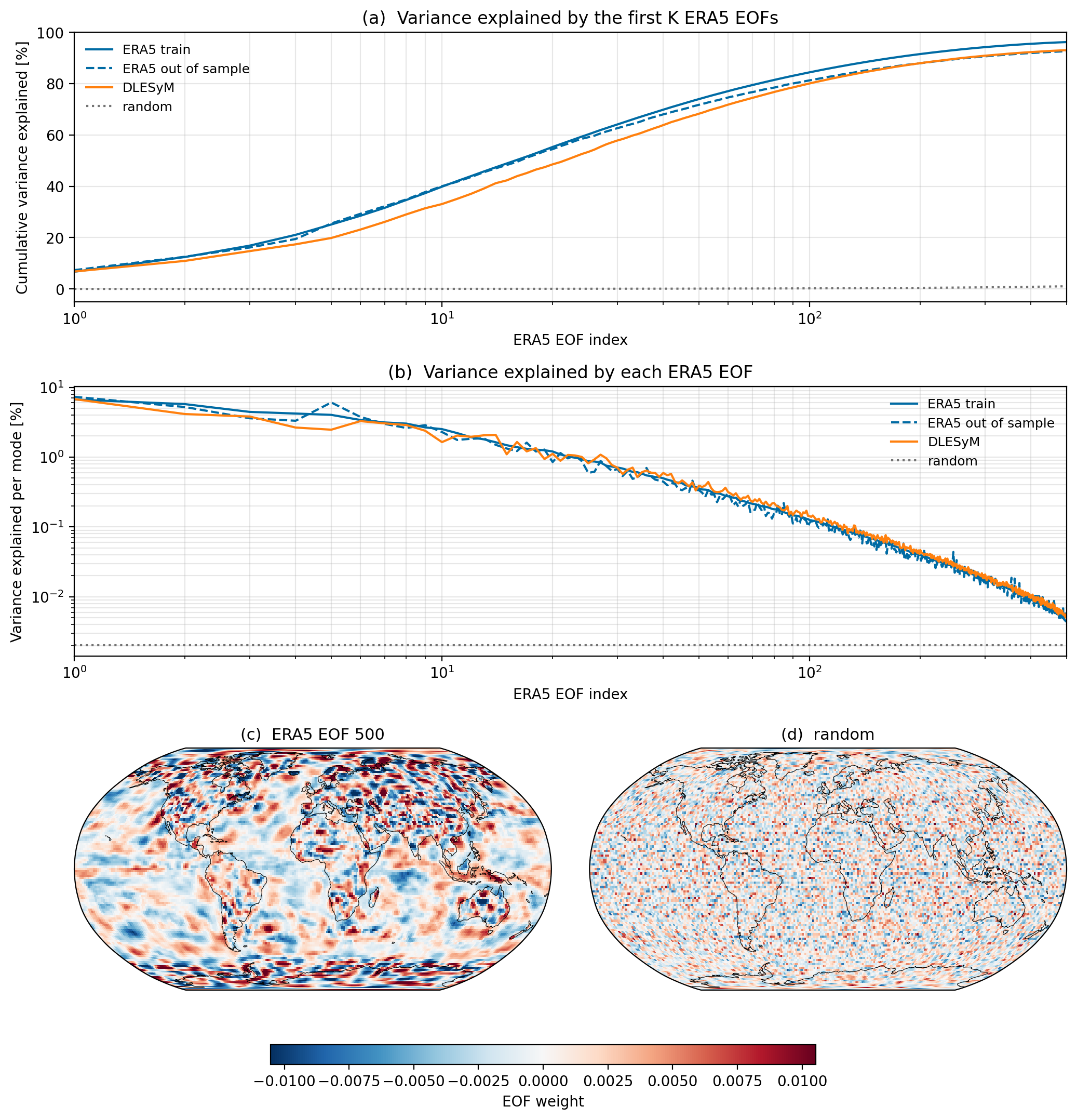}
    \caption{The ERA5 EOF basis. (a) Cumulative fraction of variance explained
    by the first $K$ ERA5 EOFs, for ERA5 training data, withheld ERA5 data, and
    DLESyM data. At $K=500$ the EOFs explain 96\% of the training variance and
    93\% of both the withheld ERA5 and the DLESyM variance. (b) Variance
    explained by each EOF. The dotted line shows the expected variance of a
    random unit vector, $1/d$. (c) ERA5 EOF 500 and (d) a random unit vector.}
    \label{fig:eof_basis_spectra}
\end{figure}

The spatial patterns of high-order EOFs estimated from the 37-year record are individually poorly determined. This sampling error predominantly mixes modes with nearby eigenvalues \cite{north1982sampling}: when the sampling error of an eigenvalue, $\delta\lambda_k \approx \lambda_k\sqrt{2/N}$ for $N$ independent samples, is comparable to its spacing from a neighboring eigenvalue, the two sample EOFs become arbitrary mixtures of the true ones, which for the ERA5-derived basis happens at mode 18. For our purposes, this means that the high-order EOFs approaching $K_{\max}$ are not individually meaningful, but for the purposes of training a skillful LIM they merely need to provide a basis that explains the predictable variance in the fields. The highest-mode EOFs considered here do not degenerate into random noise, their spatial structure follows the land--sea distribution of small-scale variance, unlike a random direction (Figure~\ref{fig:eof_basis_spectra}c,d). More pertinently, the variance they capture in independent data, both withheld ERA5 and DLESyM, remains well above that of a random basis vector out to $K_{\max}$ (Figure~\ref{fig:eof_basis_spectra}b) and tracks the spectrum of the in-sample ERA5 training data fairly closely. To reiterate, the forecast skill of the LIM depends on accurately estimating the lagged covariances within the retained subspace, not on resolving neighboring EOFs individually.

The main risk of estimating the basis from the limited ERA5 record would be if it describes the DLESyM climate poorly. However, while the $K_{\max}$ EOFs capture 96\% of the ERA5 training variance, they capture 93\% of both withheld ERA5 data and DLESyM data (Figure~\ref{fig:eof_basis_spectra}a): the reduction relative to the training data reflects in-sample fitting rather than a mismatch with the DLESyM climate. Note that the slight under-fitting of variance by the ERA5 EOFS on DLESyM data up to $K\sim100$ demonstrates that there is a non-zero domain shift transferring from DLESyM to ERA5. Ultimately, the success of $\hat{\mathbf{G}}_{\mathrm{DLESyM}}$ and $\hat{\mathbf{G}}_{\mathrm{mixed}}$ for real-world forecasting shows that the ERA5-derived basis is sufficient for the large DLESyM sample to constrain the covariances at higher $K$ than we can achieve with ERA5 alone.

\subsection{Evaluation of learned covariances through Gaussian posterior negative log-likelihood}\label{appendix:lim:nll}

To assess the full Gaussian forecast distribution, and not only its mean, we evaluate the posterior negative log-likelihood (NLL) in a common reference space spanned by the first $K_{\max}=500$ ERA5 EOFs. For a LIM with truncation $K$ and mixing weight $\alpha$, the residual for sample $i$ is
\begin{equation}
    r_i = \mathbf{E}_{K_{\max}}^{\top}y'_i
        - \boldsymbol{\Pi}_K\,\hat{\mathbf{G}}_{z}\,\mathbf{E}_K^{\top}x'_i,
\end{equation}
where $\boldsymbol{\Pi}_K$ embeds the $K$-dimensional forecast in the reference space, padding modes $K+1,\dots,K_{\max}$ with zero anomaly. The forecast-error covariance is estimated separately for each $(K,\alpha)$ from the training residuals,
\begin{equation}
    \hat{\boldsymbol{\Sigma}} = \frac{1}{n}\sum_{i=1}^{n} r_i r_i^{\top} + \epsilon\mathbf{I},
\end{equation}
with a small diagonal regularization $\epsilon=10^{-4}$. In the unresolved modes the residual is the target itself, so these retain their climatological variance.
The NLL is then evaluated on held-out samples,
\begin{equation}
    \mathrm{NLL} = \frac{1}{2}\left[K_{\max}\log 2\pi
        + \log\det\hat{\boldsymbol{\Sigma}}
        + \left\langle r^{\top}\hat{\boldsymbol{\Sigma}}^{-1}r\right\rangle_{\mathrm{val}}\right].
    \label{eq:lim_nll}
\end{equation}
This is the NLL of the conditional (posterior) forecast distribution, $\mathcal{N}(\hat{\mathbf{G}}x', \hat{\boldsymbol{\Sigma}})$ projected onto the reference space. Because every model is scored in the same $K_{\max}$-dimensional space, values are comparable across $K$. We report them relative to the NLL of a zero-anomaly climatological forecast, whose covariance is the training covariance of the target coefficients.

To see what the NLL measures beyond the forecast mean, let $\mathbf{S} = \langle r r^{\top}\rangle_{\mathrm{val}}$ denote the realized second moment of the held-out errors. Then
\begin{equation}
    \mathrm{NLL}
    = \underbrace{\frac{1}{2}\left[K_{\max}\log 2\pi + \log\det\mathbf{S} + K_{\max}\right]}_{\text{error magnitude}}
    + \underbrace{D_{\mathrm{KL}}\!\left(\mathcal{N}(0,\mathbf{S})\,\Vert\,\mathcal{N}(0,\hat{\boldsymbol{\Sigma}})\right)}_{\text{covariance calibration}}.
\end{equation} 
The first term decreases as forecast errors shrink, which is the information also contained in the MSE, although weighted logarithmically across modes. The second term vanishes only when the predicted error covariance matches the realized one, in magnitude, in its distribution across modes, and in the correlations between modes. It penalizes both over-confident forecasts, whose $\hat{\boldsymbol{\Sigma}}$ is too small, and under-confident ones. This is why ERA5-only LIMs at high truncation score poorly: their in-sample residual covariance underestimates the held-out errors, so the calibration term grows sharply. Because every mode contributes equally in logarithmic terms, the NLL is dominated by the many low-variance modes, and so complements the MSE, which is dominated by the leading modes.

\subsection{Why not use the LIM's Gaussian posterior covariance for probabilistic forecast?}\label{appendix:lim:ensemble}

Since fitting the LIM model is posed as a Gaussian maximum likelihood estimation problem, a natural choice for generating the ensemble spread for the forecast would be to assume the Gaussian posterior distribution (Equation \ref{eq:lim_Sigma_pop}). In practice, we find this to yield no major improvement in CRPS over the climatological ensemble approach.

To estimate the forecast-error covariance, we compute the covariance of the EOF-space residuals $r_i = \mathbf{E}^{\top}y'_i - \hat{\mathbf{G}}_{z}\mathbf{E}^{\top}x'_i$ over the $n$ training pairs,
\begin{equation}
    \hat{\boldsymbol{\Sigma}}_{z} = \frac{1}{n}\sum_{i=1}^{n} r_i r_i^{\top}.
    \label{eq:lim_Sigma_z}
\end{equation}
For the ERA5-only LIM this equals the estimate of Eq.~\eqref{eq:lim_Sigma_pop} in the EOF subspace; for the mixed LIM it provides the corresponding estimate. Because the forecast has no component outside the subspace, the physical-space error covariance also includes the target variance there. For scoring we require only the per-pixel variances,
\begin{equation}
    \hat{\sigma}_j^{2} = \left[\mathbf{E}\,\hat{\boldsymbol{\Sigma}}_{z}\,\mathbf{E}^{\top}\right]_{jj} + v_j,
    \qquad
    v_j = \frac{1}{n}\sum_{i=1}^{n}\left[(\mathbf{I}-\mathbf{E}\mathbf{E}^{\top})\,y'_i\right]_j^{2},
    \label{eq:lim_pixel_var}
\end{equation}
where $v_j$ is the variance of the target anomaly outside the EOF subspace at pixel $j$.

The resulting Gaussian forecast distribution at pixel $j$ is
\begin{equation}
    y'_j \sim \mathcal{N}\!\left(\hat{\mu}_j,\, \hat{\sigma}_j^{2}\right),
    \qquad
    \hat{\mu}_j = \left[\mathbf{E}\,\hat{\mathbf{G}}_{z}\,\mathbf{E}^{\top}x'\right]_j,
\end{equation}
for which the CRPS has the closed form \cite{gneiting2005calibrated}
\begin{equation}
    \mathrm{CRPS}\!\left(\mathcal{N}(\hat{\mu}_j,\hat{\sigma}_j^{2}),\, y'_j\right)
    = \hat{\sigma}_j\left[\, z_j\left(2\Phi(z_j)-1\right) + 2\varphi(z_j) - \frac{1}{\sqrt{\pi}} \,\right],
    \qquad
    z_j = \frac{y'_j - \hat{\mu}_j}{\hat{\sigma}_j},
    \label{eq:gaussian_crps}
\end{equation}
where $\Phi$ and $\varphi$ are the standard normal cumulative distribution and density functions.

\begin{figure}[t]
    \centering
    \includegraphics[width=\columnwidth]{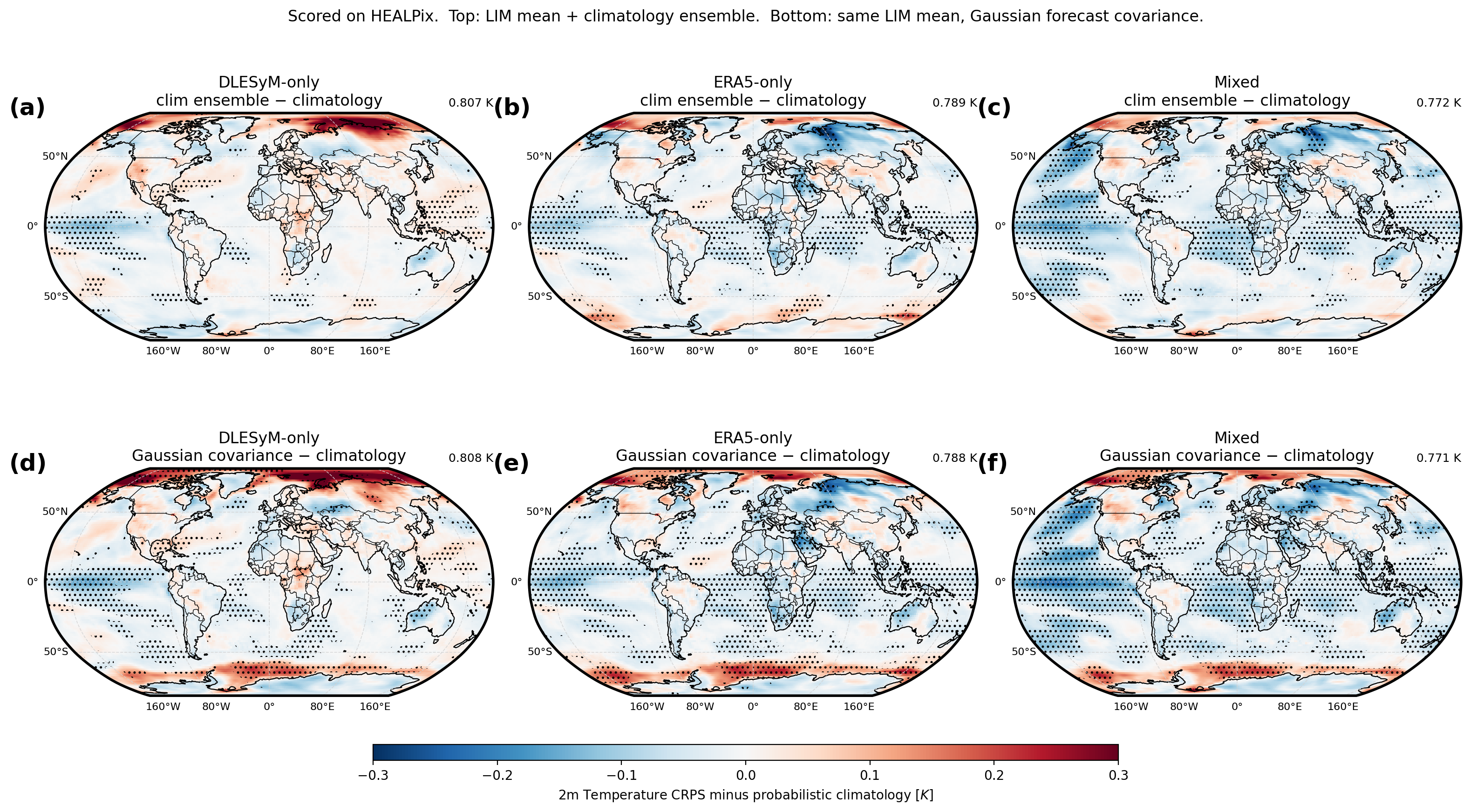}
    \caption{Effect of the LIM ensemble construction on real-world week-4
    2\,m temperature CRPS. Each panel shows the CRPS difference from the
    probabilistic climatology (blue indicates improvement), for the
    DLESyM-only (left), ERA5-only (centre), and mixed (right) LIMs, all with
    the same forecast mean. (a--c) Ensemble formed by adding climatological
    anomalies to the LIM mean, as used in the main text. (d--f) Gaussian
    forecast distribution using the LIM error covariance
    (Eq.~\ref{eq:lim_pixel_var}). The number at the top right of each panel is
    the global mean CRPS. Stippling indicates differences that are
    statistically significant (Appendix~A). Scores are computed on the HEALPix
    grid.}
    \label{fig:lim_gaussian_vs_clim_ensemble}
\end{figure}

Comparing the CRPS obtained using this Gaussian forecast distribution with that obtained using the climatological ensemble of the main text, the two approaches give very similar global mean CRPS (Figure~\ref{fig:lim_gaussian_vs_clim_ensemble}, panels a--c against d--f). Regionally, the Gaussian distribution improves CRPS somewhat over the tropical oceans, where the LIM is most skillful and the climatological spread is too wide. It degrades CRPS at high latitudes, potentially indicating strongly seasonal or non-Gaussian variability that is poorly represented by a single all-season Gaussian covariance but is captured naturally by climatological anomalies drawn from the same calendar dates. For simplicity, we use the climatological ensemble in the main text. This matches the construction of the probabilistic climatology baseline, so that CRPS differences reflect the skill of the forecast mean, and it is straightforward to score on the latitude--longitude grid used for the other forecasts, since each ensemble member can simply be regridded. Regridding the Gaussian distribution, by contrast, would require the error correlations between neighboring pixels, not just the per-pixel variances of Eq.~\eqref{eq:lim_pixel_var}. The comparison in Figure~\ref{fig:lim_gaussian_vs_clim_ensemble} was thus performed on the HEALPix grid on which the LIM is defined, whereas all scoring in the main text is performed on the latitude--longitude grid.

Both constructions considered here yield a forecast spread that is independent of the initial state: the LIM error covariance does not depend on $x'$, and neither do the climatological anomalies. Since S2S predictability is likely state-dependent, a more sophisticated state-dependent spread may further improve the probabilistic skill of the LIM, but this is beyond the scope of this study.

\section{Damped Persistence}\label{appendix:baselines}

To provide an additional statistical baseline for real-world S2S forecasting in addition to climatology, we fit a simple damped persistence model to the ERA5 training data and evaluate S2S forecast performance over the same initialization dates as considered for the other methods. Below we describe the implementation of this baseline.

\subsection{Damped Persistence}

As a classical statistical baseline, we use a local damped-persistence model of
ERA5 anomalies. Let $x$ denote the initial state, defined as the mean of the seven daily fields ending at the forecast initialization time, and let $y$ denote the verification target, defined as the temporal mean over the same lead and averaging window used by the other S2S forecasts (a 7-day mean centered 28~days after initialization). Both fields are represented on the HEALPix grid and treated independently at each grid cell and for each variable.

We remove the seasonal cycle using daily climatology $\mathrm{clim}$, constructed from historical ERA5 and smoothed with a 7-day running mean. The initial and target climatologies, $\mathrm{clim}_{\mathrm{in}}$ and $\mathrm{clim}_{\mathrm{tgt}}$, are retrieved for the input and target windows, respectively. Anomalies are then
\begin{equation}
x' = x - \mathrm{clim}_{\mathrm{in}},\qquad y' = y - \mathrm{clim}_{\mathrm{tgt}}.
\end{equation}

At each grid cell we fit a through-origin linear regression of target anomalies on initial anomalies,
\begin{equation}
\hat{y}' = a\,x',
\end{equation}
with
\begin{equation}
a=\frac{\sum_n x'_n y'_n}{\sum_n (x'_n)^2},
\end{equation}
where the sum runs over training samples from the ERA5 training period. The residual scale is estimated as
\begin{equation}
b = \sqrt{\mathbb{E}\!\left[(y' - a x')^2\right]}.
\end{equation}
The corresponding physical-space forecast is therefore a damped persistence of the initial anomaly about the target climatology,
\begin{equation}
\hat{y} = \mathrm{clim}_{\mathrm{tgt}} + a\,(x - \mathrm{clim}_{\mathrm{in}}).
\end{equation}
To form a probabilistic ensemble, we add independent Gaussian residual noise at each grid cell,
\begin{equation}
y^{(m)} = \mathrm{clim}_{\mathrm{tgt}} + a\,(x - \mathrm{clim}_{\mathrm{in}}) + b\,z^{(m)}, \qquad z^{(m)}\sim\mathcal{N}(0,1),
\end{equation}
for ensemble members $m=1,\ldots,M$.
We fit separate coefficients $(a,b)$ for each calendar month of the initialization date to allow the model to vary seasonally.

\section{Early Exploration of Quantile Regression Approach}

In this manuscript we presented results from formulating our distilled student model as a conditional diffusion model. In early explorations for this project we explored an alternative approach where we used quantile regression instead of conditional diffusion for the probabilistic forecasting. The experiments conducted using this method were focused on quickly exploring different methods within our compute budget rather than systematic ablations. Below we include informal observations and our interpretations from these explorations in the hope they may benefit future researchers.

\subsection{Quantile Regression Formulation}

To obtain a probabilistic forecast in a single model timestep using quantile regression, we discretized our forecast into climatological quintile bins from the preceding 20 years and tasked a neural network with directly forecasting the quintile probabilities rather than generating ensemble member forecasts. We used the same UNet backbone architecture as was presented in the manuscript but removed the $\sigma$ level conditioning and used a cross-entropy loss function to classify which quintile bin the target falls in. These experiments were primarily focused on S2S forecasting.

\subsection{Findings}

\textbf{Observation 1.} Training the quantile regression method on 11,000 years of DLESyM simulation data for S2S forecasting led to very weak learning, with slowly converging loss curves even after exploring a wide range of hyperparameters.

\textbf{Interpretation.} S2S forecasting is a challenging task with very limited prediction skill. The quantile regression framing with a cross-entropy loss function has the seemingly desirable property that a randomly initialized network starts with the same skill as probabilistic climatology (predicting equal quintile probabilities). However, in a weak predictability regime like S2S this likely makes it challenging for the network to learn powerful representations since it gets only very weak signals from the loss function.

\textbf{Observation 2.} Using curriculum learning \cite{bengio2009curriculum} on lead times by fine-tuning a network initially trained for 7-day lead time quantile regression on S2S lead times improves skill beyond training from scratch at S2S. Training on an equal mix of different lead times achieves a similar effect.

\textbf{Interpretation.} This supports our interpretation of observation 1 that the S2S forecasting challenge provides too challenging an initial task for the randomly initialized network to learn meaningful data representations.

\textbf{Observation 3.} Training a S2S quantile regression model from scratch on ERA5 achieves better skill than the mdoel trained on 11,000 years of DLESyM data on withheld DLESyM data. Further, fine-tuning the DLESyM pre-trained quantile regression model on ERA5 leads to the same performance as training from scratch on ERA5 but does not improve over training from scratch.

\textbf{Interpretation.} Training on a smaller dataset allows the network to overfit, without the implicit regularization of training on 11,000 years of simulation. This leads to some initial skill on the validation set compared to the network trained on 11,000 years which should be thought of as an almost randomly initialized network due to its weak learning. But the network quickly overfits and can't be improved beyond a certain point using the limited ERA5 dataset. Fine-tuning the DLESyM pre-trained network on ERA5 essentially reverts to training from scratch on ERA5 since the network was unable to learn meaningful representations on the large DLESyM dataset.

\textbf{Observation 4.} Despite being presented with the same training dataset and task, conditional diffusion model shows real-world benefit of pre-training on DLESyM and performance scaling with increasing synthetic data.

\textbf{Interpretation.} The diffusion objective allows the network to learn meaningful data representations even in challenging predictability regimes since it focuses on gradually de-noising corrupted examples from the training set. This pushes the network past the early training problems observed for quantile regression and allows it to properly exploit the large simulation dataset.

\section*{Open Research}

Our code for long-range distillation will be released in a public GitHub repository upon publication. The DLESyM teacher model code used in this manuscript is available in the NVIDIA Earth2Studio package \cite{e2s}.

\acknowledgments

S.M. thanks Peter Harrington, David Pruitt, and Nathaniel Cresswell-Clay for discussions. We thank the two anonymous reviewers for thoughtful reviews which improved the manuscript.
    
\clearpage

\bibliography{references}

\end{document}